\documentclass{article}

\usepackage[square,numbers,sort&compress]{natbib}
\usepackage[final]{format}

\usepackage{multicol}
\usepackage{times}
\usepackage{cancel}
\usepackage{color}
\usepackage{caption}
\usepackage{subcaption}
\usepackage{graphicx}
\usepackage{inconsolata}
\usepackage{xcolor}
\usepackage{float}
\usepackage{ctable}
\usepackage{boldline}
\usepackage{cancel}
\usepackage{soul}
\usepackage{wrapfig}
\usepackage{minibox}
\usepackage{appendix}
\usepackage{multirow}

\usepackage[ruled, noline]{algorithm2e}
\SetKwFor{ForPar}{for}{do in parallel}{end forpar}

\usepackage[T1]{fontenc} 
\usepackage[latin1]{inputenc}
\usepackage{babel}

\usepackage{amsmath, amsthm, amssymb}
\usepackage[capitalise]{cleveref} 
\usepackage{textcomp}
\usepackage{stmaryrd}
\usepackage{upgreek}
\usepackage{bm}
\usepackage{bbm}
\usepackage{url}
\usepackage{breakurl}
\usepackage{xspace}
\usepackage{afterpage}

\newcommand{\deriv}[2]{\frac{d #1}{d #2}}

\newcommand{\expect}[2]{\mathbb{E}_{#1}\Big[ #2 \Big]}

\newcolumntype{L}[1]{>{\raggedright\arraybackslash}p{#1}}
\newcolumntype{C}[1]{>{\centering\arraybackslash}p{#1}}
\newcolumntype{R}[1]{>{\raggedleft\arraybackslash}p{#1}}

\def\L{\mathcal{L}}
\def\real{\mathbb{R}}

\def\L{\mathcal{L}}

\def\O{\mathcal{O_{\tau}}}

\def\xb{X}
\def\ub{U}

\def\x{\mathbf{x}}
\def\u{\mathbf{u}}
\def\w{\mathbf{w}}

\def\cost{C}

\def\pib{\mathrm{\boldsymbol{\pi}}}
\def\thetab{\mathrm{\boldsymbol{\theta}}}

\newcommand{\hide}[1]{}
\DeclareMathOperator*{\argmax}{arg\,max} 
\DeclareMathOperator*{\argmin}{arg\,min}
\newcommand{\norm}[1]{\big|\big| #1 \big|\big|}

\DeclareMathAlphabet\mathbfcal{OMS}{cmsy}{b}{n}

\title{
Stein Variational Model Predictive Control
}

\author{Alexander Lambert$^{1, 2}$ \thanks{E-mail: {\tt \small alambert6@gatech.edu}} \quad Adam Fishman$^{2, 3}$ \quad Dieter Fox$^{2, 3}$\\[5pt]
\bf{Byron Boots}$^{2, 3}$ \quad \bf{Fabio Ramos}$^{2, 4}$ \\[5pt]
$^{1}$ Georgia Tech \quad $^{2}$ NVIDIA \quad $^{3}$ University of Washington \quad $^{4}$ The University of Sydney
}

\begin{document}
\maketitle
\vspace{-20pt}
\begin{abstract}
Decision making under uncertainty is critical to real-world, autonomous systems. Model Predictive Control (MPC) methods have demonstrated favorable performance in practice, but remain limited when dealing with complex probability distributions. In this paper, we propose a generalization of MPC that represents a multitude of solutions as posterior distributions. By casting MPC as a Bayesian inference problem, we employ variational methods for posterior computation, naturally encoding the complexity and multi-modality of the decision making problem. We present a Stein variational gradient descent method to estimate the posterior directly over control parameters, given a cost function and observed state trajectories. We show that this framework leads to successful planning in challenging, non-convex optimal control problems. The supplementary video can be viewed here: \url{https://youtu.be/pgBCii1NfA0}

\end{abstract}


\section{Introduction}
\label{sec:introduction}
Model predictive control (MPC) is a powerful framework for sequential decision making in robotics~\cite{infotheoreticmpc, dmd-mpc}. This success can be largely attributed to its simplicity and scalability in dealing with stochastic,  non-stationary optimization problems encountered on real systems. MPC has been applied effectively in different areas, including autonomous driving~\cite{williams2016aggressive,infotheoreticmpc, dmd-mpc}, humanoid locomotion~\cite{kasaei2019robust}, and dextrous manipulation~\cite{kumar2016optimal}. However, common approaches to MPC often fall short in their ability to adequately contend with complex, multi-modal distributions over possible actions. For instance, such distributions may arise from non-convexity of constraints, such as obstacles~\cite{kappen2005path, gomez2016real} or from multiple goal locations~\cite{wiegerinck2012stochastic}. Although common sampling-based SOC algorithms have been shown to exhibit symmetry-breaking in the presence of sudden disturbances~\cite{williams2018robust} and multiple optima~\cite{kappen2005linear}, the sampling scheme may inadequately resolve the true posterior~\cite{kappen2005path}. We require a new class of MPC algorithms that can effectively contend with complex, non-Gaussian distributions.

In the following, we formulate MPC as a Bayesian inference problem, where the target posterior is defined directly over control policy parameters or control inputs, as opposed to joint probabilities over states and actions~\cite{rawlik2013, levine_tutorial}. By taking this perspective, we can construct a relative-entropy minimization problem to approximate the posterior, and leverage recent advances in variational inference~\cite{Jordan99, Beal03, Wainwright2008} to derive the optimal distribution over parameters. Specifically, we use Stein variational gradient descent (SVGD)~\cite{Liu16}, to infer a set of solutions which constitute a nonparametric approximation to the posterior distribution over decision parameters, given state observations and a defined cost function. The generality of this approach offers flexibility in designing appropriate MPC algorithms, and is closely related to common MPC approaches. Additionally, we show how this framework can be extended to general trajectory optimization problems. 

\textbf{Related Work \quad}\label{sec:related_work} The duality between probabilistic inference and optimization for stochastic optimal control (SOC) has been examined extensively in previous work \cite{todorov2008duality, rawlik2010approximate,toussaint2009,kappen2012optimal}. A wide range of approximate inference methods have been proposed, including  Expectation-Propagation (EP) and moment-matching approaches to control and trajectory optimization \cite{williams2016aggressive, toussaint2009}.  These methods typically assume a restricted form of the target posterior distribution over controls (usually in the exponential family), which, under simplifying assumptions, return the optimal control input. However, these distributions are often insufficiently expressive for many SOC problems, where non-convexity may arise from nonlinear dynamics or non-convex cost functions, for instance. 
Our interpretation of MPC as an approximate inference problem is perhaps most closely related to the formulation presented in \cite{rawlik2010approximate,rawlik2013}. Here, an iterative KL-minimization problem for finite-horizon problems is presented, where the prior is defined to be the control distribution obtained from the previous iteration. However, this does not include a strategy for contending with non-stationary distributions, where the posterior may change between iterations, and the analysis is restricted to exact inference of the log-partition function.

Expectation-Maximization (EM) approaches to SOC have also been considered
\cite{toussaint2006probabilistic, Okada2019VariationalIM, watson2020stochastic}, which iteratively optimize a variational lower-bound. However, these approaches share the limitations inherent with common EM strategies, in that the form of the posterior distribution is assumed to be known and tractable. Additionally, representations based on mixture models must contend with  mode-collapse and poor local minima, especially prevalent in higher dimensions~\cite{bishop2006pattern, ormoneit1998averaging}. These issues can be mitigated, for example, by introducing entropy regularization heuristics~\cite{Okada2019VariationalIM}.

The Path Integral (PI) formulation of stochastic optimal control \cite{theodorou2010generalized, theodorou2012duality, theodorou2015nonlinear} bears close resemblance to the open-loop-controls characterization of our proposed approach, where a particular choice of the marginal log-likelihood is assumed to contain an exponentiated cost. In fact, PI attempts to minimize a variational lower-bound with respect to the controlled stochastic dynamics. This follows from the perspective presented here. This comparison applies equally to KL-control~\cite{kappen2012optimal}, the discrete-time counterpart to PI-control. A more detailed discussion is included in appendix~\ref{app:pi_connection}.

The application of SVGD for approximate inference in decision making problems has been previously explored in the context of maximum entropy reinforcement learning, where policies are trained using collected experience and updated offline. In \cite{Liu2017SteinVP}, the network-based policy is represented using a set of Stein particles to generate a single-step action. However, it is unclear how to appropriately design kernels and evaluate them efficiently in order to counter the effects of increasing dimensionality (\textit{i.e.} network size) using this approach. By contrast, the soft Q-learning algorithm~\cite{haarnoja2017reinforcement} evaluates the SVGD gradient over single-step actions. This is then subsequently backpropagated through the policy to update the parameters. To our knowledge, the non-parametric variational inference approach of \cite{Liu16} has yet to be applied to MPC, or general SOC and planning problems. The dependence on length of the planning horizon poses interesting challenges for scaling and sample-efficient computation. Furthermore, the proposed approach is developed with the online setting in mind, and addresses the problem of rapidly-changing, non-stationary target distributions. Lastly, we make explicit connections to existing SOC methods from a theoretical standpoint. 
\vspace{-5pt}

\section{Model Predictive Control}\vspace{-5pt}
\label{sec:MPC}

We consider the discrete-time stochastic dynamical system: $\x_{t+1} \sim  f\left(\x_t, \u_t \right)$, where at time $t$, the system state is denoted by $\x_t \in \real^n$ and the control input as $\u_t \in \real^d$. The stochastic transition map $f : \real^n \times \real^d \rightarrow \real^n$ randomly produces the subsequent state $\x_{t+1}$ and this state is accompanied by an instantaneous cost $c(\x_t, \u_t)$.

Over a time horizon $H$,  we define a control trajectory as a sequence of control inputs beginning at time $t$ : $\ub_t\triangleq\left( \u_t, \u_{t+1}, ..., \u_{t+H-1}\right)$. Similarly, we define the state trajectory: $\xb_t \triangleq \left( \x_t, \x_{t+1}, ..., \x_{t+H-1}, \x_{t+H}\right)$. 
The total cost incurred over $H$ timesteps can then be specified as
\begin{align}\label{eq:cost}
C(\xb_t, \ub_t) =  c_{\mathrm{term}}(\x_{t+H}) + \sum_{h=0}^{H-1} c(\x_{t+h}, \u_{t+h}), 
\end{align}
where $c_{\mathrm{term}}(\cdot)$ is the terminal cost. As in \cite{dmd-mpc}, we define an instantaneous feedback policy, $\pi_{\theta_t}(\x_t)$, as a parameterized probability distribution $p(\u_t | \x_t;\ \theta_t)$ used to generate a control input at  time $t$ given the $\x_t$, \textit{i.e.},  $\u_t \sim \pi_{\theta_t}(\x_t)$, where $\theta_t \in \Theta$, is the set of feasible parameter values. MPC describes the process of finding the optimal, time-indexed sequence of policy parameters $\thetab_t \triangleq \left(\theta_t, \theta_{t+1}, ..., \theta_{t+H-1}\right) $, which determine the sequence of instantaneous feedback policies $\pib_{\thetab_t}\triangleq \left( \pi_{\theta_t}, \pi_{\theta_{t+1}}, ..., \pi_{\theta_{t+H-1}}\right)$.
At each time step, we must find $\thetab_t$, the parameters that define the optimal policy. We can do this by defining a statistic $J(\cdot)$ on cost $C(\xb_t, \ub_t) $ where the minimal $J(\cdot)$ occurs at the optimal $\thetab_t$.
In real-world situations, the true dynamics function $f$ is often unavailable, and is commonly estimated using a parameterized function $\hat{f}_\xi$ with parameters $\xi$.  As such, we define the surrogate loss function $\hat{J}(\pib_\thetab;  \x_t)=\expect{\pib_\thetab, \hat{f}_\xi} {\cost(\xb_t, \ub_t)}$.  For each MPC-step, the optimal decision is defined as $\thetab_t = \argmin_{\thetab}\ \hat{J}(\pib_\thetab; \x_t)$ which parameterizes the optimal policy $\pi_{\theta_t}$, from which we can sample a new control value for the first timestep: $\u_t \sim \pi_{\theta_t}(\x_t)=p(\u_t|\x_t; \theta_t)$. This is then executed on the physical system to generate the next state value: $\x_{t+1} \sim f (\x_t, \u_t)$.

\section{Bayesian Model Predictive Control}\vspace{-5pt}
\label{sec:bayes}

\textbf{MPC as Bayesian Inference \quad}\label{sec:bayes_mpc} Optimal control can be framed as Bayesian inference by considering the distribution over parameters $\thetab$. Similarly to \cite{rawlik2010approximate, levine_tutorial}, we introduce an auxiliary binary random variable $\O\in\{0,1\}$ to indicate optimality of the state-action trajectory $\tau = (\xb_t, \ub_t)$ with respect to the cost function $C(\cdot)$. Using Bayes' rule, the distribution of parameters $\thetab$ conditioned on the requirement for optimal trajectories ($\O=1$) and the current state $\x_t$ can be expressed as
\begin{align}\label{eqn:bayes}
p_t(\thetab | \O=1; \xi, \x_t)  &=  \frac{p_t(\O=1 |  \thetab; \xi, \x_t)\ p_t(\thetab; \x_t)} {\int p_t(\O=1 | \thetab; \xi, \x_t)\ p_t(\thetab; \x_t)\  d\thetab},
\end{align}
where explicit dependence on state $\x_t$ is included for generality. In the remaining discussion, we will denote $\O=1$ simply as $\O$ without ambiguity. The likelihood $p_t(\O|\thetab; \xi,\x_t)$ is defined as the marginal probability over all possible control and state trajectories: 
\begin{align}\label{eq:llh_func}
p_t(\O|\thetab; \xi,\x_t)  &= 
\int\int p(\O|\xb_t, \ub_t)\ 
p(\xb_t, \ub_t \ | \thetab;\xi,\x_t)\ d \ub_t\ d\xb_t
\end{align}
where  $p(\O|\xb_t, \ub_t)$ is the probability of optimality given the observed trajectory $\tau = (\xb_t, \ub_t)$, and $p(\xb_t,\ub_t|\thetab; \xi, \x_t)$ is the joint probability of state-control trajectories, conditioned on parameters $\thetab$ and assumed dynamics model $\hat{f}_\xi =p_\xi (\x_{t+1}|\x_{t}, \u_{t})$. In the discrete-time case, the joint probability can be factorized as
\begin{align}
p(\xb_t,\ub_t | \thetab;  \xi,  \x_t) = 
\prod_{h=0}^{H-1} p_\xi(\x_{t+h+1}|\x_{t+h}, \u_{t+h}) 
\pi_{\theta_{h}}(\x_{t+h}),
\end{align}
where the current state $\x_t$ is observed. If dynamics parameters must also be inferred, the equations can be extended to include $\xi$ as an inference variable, where the posterior is defined over both parameters: $p_t(\thetab, \xi | \O; \x_{t}) \propto p_t(\O|\thetab, \xi; \x_{t} ) \, p_t(\thetab | \xi; \x_{t})\, p(\xi)$, suggesting the definition of an alternative latent random variable, such as $\mathbf{\Theta} = \left(\thetab, \xi \right)$. 

We can further model $p(\O|\xb_t, \ub_t)$ using a non-negative function $\L(\tau) \propto p(\O|\xb_t, \ub_t)$, which we refer to as the ``cost-likelihood". This is defined to be the composition $\L = g \circ \cost $ of the cost function $\cost(\cdot)$ with a monotonically decreasing function $g(\cdot)$. The likelihood in \eqref{eq:llh_func} then takes the form
\begin{align}\label{eq:llh_func_2}
p_t(\O|\thetab; \xi,\x_t) &\propto \int \mathcal{L}(\tau)\ p(\tau \ | \thetab;\xi,\x_t)\ d \tau 
= \expect{\pib_{\thetab}, \hat{f}_\xi}{\ \mathcal{L}(\tau)} \ .
\end{align}

\textbf{Nonparametric Bayesian MPC \quad}\label{sec:nonparam_mpc} Instead of performing inference over policy parameters, the Bayesian MPC formulation can be used for inference over control input sequences. In this case, the inference variable $\thetab$ is defined as the sequence of open-loop controls: $\thetab \triangleq  \left( \u_t, \u_{t+1}, ..., \u_{t+H-1}\right)$. This can be interpreted as a nonparametric version of Bayesian-MPC, as no assumption is made on the existence of a parametrized policy $\pib$.  Although the relationship in \eqref{eqn:bayes} still holds, the likelihood function $p_t(\O |  \thetab;  \xi, \x_t)$ must then be re-defined as
\begin{align}\label{eqn:nonparam_likelihood}
p_t(\O |  \thetab;  \xi, \x_t) =
\int p(\O | \xb_t, \thetab)\ 
p(\xb_t | \thetab; \xi, \x_t)\ d\xb_t \ \propto \ \expect{ \hat{f}_\xi}{\ \mathcal{L}(\xb_t, \thetab)} \ ,
\end{align}
where  $p(\xb_t| \thetab;  \xi, \x_t) $ is the probability of state trajectories, conditioned on decisions $\thetab$ and assumed dynamics model $\hat{f}_\xi =p_\xi (\x_{t+1}|\x_{t}, \u_{t})$.  In the discrete-time case, this can be written as the product of state transition probabilities along a trajectory:
\begin{align}
p(\xb_t  | \thetab;  \xi,  \x_t) = 
\prod_{h=0}^{H-1} p_\xi(\x_{t+h+1}|\x_{t+h}, \theta_{t+h}) \ .
\end{align}
Furthermore, if we assume a fixed current state ($\x_t=\textrm{const.}$), we can apply this model to trajectory optimization problems by performing approximate inference on the posterior $p(\thetab | \O; \xi)$, and taking the {\em maximum a posteriori} estimate, $\thetab^* = \argmax_\thetab\ p(\thetab | \O; \xi)$.

\section{Inference for Bayesian MPC}\vspace{-5pt}
\label{sec: inference}

\textbf{Variational Inference \quad} Variational inference poses posterior estimation as an optimization task where a candidate distribution $q^*(\thetab)$ within a distribution family $\mathcal{Q} = \{ q(\thetab) \}$ is chosen to best approximate the target distribution $p(\thetab | \O)$. This is typically obtained by minimizing the Kullback-Leibler (KL) divergence:
\begin{align}\label{eq:kl_min}
q^* &= \argmin_{q\in \mathcal{Q}}\ D_{KL} \left( q(\thetab) ||  p(\thetab | \O) \right).
\end{align}
The solution also maximizes the Evidence Lower Bound (ELBO), as expressed by the following objective (details in appendix \ref{app:vi_objective}):
\begin{align}\label{eq:kl_min_2}
q^* = \argmin_{q\in \mathcal{Q}}\ -\expect{q}{\log p(\O | \thetab)}  + D_{KL} \left( q(\thetab)\ ||\ p(\thetab)\right).  
\end{align}
This optimization seeks to maximize the log-likelihood of the observations with the first term while penalizing for differences between the target and the prior with the second term. For a high-capacity model space $\mathcal{Q}$ that includes the target distribution, the second term becomes increasingly small. Selecting a model space $\mathcal{Q}$ with both high capacity and computational efficiency is critical to variational inference, but remains a challenging problem.

\textbf{Stein Variational Gradient Descent \quad} In order to circumvent the challenge of determining an appropriate $\mathcal{Q}$, while also addressing \eqref{eq:kl_min}, we develop an algorithm based on Stein variational gradient descent (SVGD) for Bayesian inference. The nonparametric nature of SVGD is advantageous as it removes the need for assumptions on restricted parametric families for $q$. This approach approximates a posterior $p(\thetab | x)$  with a set of particles $\{\thetab^i\}_{i=1}^m$, $\thetab^i \in \real^p$. The particles are iteratively updated according to
\begin{align}
\thetab^i \leftarrow \thetab^i + \epsilon \bm{\phi}^*(\thetab^i)
\label{eq: stein_update}
\end{align}
given a step-size $\epsilon$. The function $\bm{\phi}^*(\cdot)$ lies in the unit-ball of an $\real^p$-valued reproducing kernel Hilbert space (RKHS) of the form $\mathcal{H} =\mathcal{H}_0\times ... \mathcal{H}_0 $, where $\mathcal{H}_0$ is a scalar-valued RKHS with kernel $k(\thetab', \thetab)$. It characterizes the optimal perturbation or velocity field (i.e. gradient direction) which maximally decreases the KL-divergence:
\begin{align}
\bm{\phi}^*  = \argmax_{\bm{\phi} \in \mathcal{H}} \Big\{ - \nabla_\epsilon D_{KL} \left(q_{\left[\epsilon \bm{\phi} \right]}||p(\thetab | x) \right)\,\mathrm{s.t.}\, \norm{\bm{\phi}}_{\mathcal{H}} \leq 1\Big\},
\end{align}
where $q_{\left[\epsilon \bm{\phi} \right]}$  indicates the particle distribution resulting from taking an update step $\thetab = \thetab + \epsilon \bm{\phi}(\thetab)$. This has been shown to yield a closed-form solution~\cite{Liu16} which can be interpreted as a functional gradient in RKHS and approximated with the set of particles:
\begin{align}
\hat{\bm{\phi}}^*(\thetab) = \frac{1}{m}\sum_{j=1}^m
\Big[
k(\thetab^j, \thetab)\nabla_{\thetab^j}\log p(\thetab^j || x)
+ \nabla_{\thetab^j} k(\thetab^j, \thetab) \Big].
\label{eq:phi_hat}
\end{align}
Eq.~\ref{eq:phi_hat} has two terms that control different aspects of the algorithm. The first term is essentially a scaled gradient of the log-likelihood over the posterior's particle approximation. The second term is known as the {\em repulsive force}. Intuitively, it pushes particles apart when they get too close to each other and prevents them from collapsing into a single mode. This allows the method to approximate complex, possibly multi-modal posteriors in MPC. When there is only a single particle, the method reduces to a standard optimization of the log-likelihood or a MAP estimate of the posterior as the repulsive force term vanishes, \textit{i.e.} $\nabla_{\thetab} k(\thetab, \thetab) = 0$. SVGD's optimization structure empirically provides better particle efficiency than other popular sampling procedures, such as Markov Chain Monte Carlo~\cite{Chen2019SteinPM}.\vspace{-3pt}

\section{Stein Variational MPC}\vspace{-5pt}
\label{sec: stein_mpc}
In this section we present our novel method for Stein inference, specifically designed around MPC requirements. The full algorithm can be found in appendix~\ref{app:algorithm}.

\textbf{Posterior Sequential Updates \quad} As described in Section~\ref{sec:bayes}, the Bayesian interpretation of MPC seeks to find the posterior distribution over decision parameters at time $t$. Recalling Eq.~\ref{eqn:bayes}:
\begin{align}\label{eqn:updated_bayes}
p_t(\thetab | \O; \xi, \x_t)  &=  \frac{p_t(\O |  \thetab; \xi, \x_t) \ \Tilde{q}_t(\thetab; \x_t)} {\int p_t(\O |  \thetab; \xi, \x_t)\ \Tilde{q}_t(\thetab; \x_t)\  d\thetab}  \quad.
\end{align}
with a prior $\Tilde{q}_t(\thetab; \x_t)$. Our approach approximates the posterior over decision parameters using a \textit{weighted} set of particles $\{\thetab^i\}^m_{i=1}$, where the proposal $q$  is defined as the empirical distribution $q(\thetab) = \sum_{i=1}^m w^i\delta(\thetab^i)$ with weights evaluated according to:
\begin{align}\label{eq:weight}
    w^i &= \frac{p_t(\O | \thetab^i; \xi, \x_t)\ \Tilde{q}_t(\thetab; \x_t)} {\sum_{j=1}^m p_t(\O | \thetab^j; \xi,  \x_t)\ \Tilde{q}_t(\thetab; \x_t)}
\end{align}
such that $\sum_{i=1}^m w^i = 1$. Following the procedure outlined in \eqref{eq: stein_update}-\eqref{eq:phi_hat}, an SVGD update can be computed for the individual particles by computing the functional gradient
\begin{align}\label{eq:svgd_mpc}
    \hat{\bm{\phi}}^*(\thetab^i) = \frac{1}{m}\sum_{j=1}^m \Big[k(\thetab^j, \thetab^i)\nabla_{\thetab^j}\log p_t( \thetab^j| \O; \xi,\x_t)
    + \nabla_{\thetab^j} k(\thetab^j, \thetab^i)\Big]
\end{align}
and performing the gradient step: $\thetab^i \leftarrow \thetab^i + \epsilon \hat{\bm{\phi}}^*(\thetab^i)$ . The evaluation of \eqref{eq:svgd_mpc} requires computation of the log-posterior gradient, which can be written as the sum of the gradients of both the log-prior and log-likelihood:
\begin{align}\label{eq:dlog_p}
\nabla_{\thetab^i}\log p_t(\thetab^i|\O; \xi,\x_t) &= \nabla_{\thetab^i} \log p_t(\O|\thetab^i;\xi,\x_t) + \nabla_{\thetab^i} \log \Tilde{q}_t(\thetab^i; \x_t) \\
&= \nabla_{\thetab^i} \log \expect{\pib_{\thetab^i}, \hat{f}_\xi}{\ \mathcal{L}(\tau)}  + \nabla_{\thetab^i} \log \Tilde{q}_t(\thetab^i; \x_t)
\end{align}

The first RHS term requires that we define the cost-likelihood function $\L$. The SV-MPC framework allows for different possible definitions for $\L$. We will consider two in particular:
\begin{align}\label{eq:dlog_p}
\bullet \quad &\textrm{Exponentiated Utility (EU):} \quad \L(\tau) = \exp \left( -\alpha C(X_t, U_t)\right), \quad \textrm{where  } \alpha > 0 \\
\bullet \quad &\textrm{Probability of Low Cost (PLC):} \quad \L(\tau) = \mathbbm{1}_{\cost \leq \cost_{\max}}(\cost(\xb_t, \ub_t))
\end{align}

It is generally assumed that the cost function $C(\xb_t, \ub_t)$ is non-differentiable with respect to the decision parameter $\thetab$, and that resulting expectations are difficult to evaluate analytically. As such, the gradients can be estimated via Monte-Carlo sampling, where a set of $N$ control and state trajectory samples are drawn from the policy and the modeled dynamics: $\{\tau^s\}_{s=1}^N \sim p(\xb_t,\ub_t | \thetab;  \xi,  \x_t)$, where $\tau^s = (\xb_t^s, \ub_t^s)$. This leads to the following approximation:
\begin{align}\label{eq:llh_approx}
\nabla_{\thetab^i} \log \expect{\pib_{\thetab^i}, \hat{f}_\xi}{\ \L(\tau)} = \frac{\expect{\pib_{\thetab^i}, \hat{f}_\xi}{\L(\tau) \nabla_{\thetab^i} \log  \pib_{\thetab^i}}}{\expect{\pib_{\thetab^i}, \hat{f}_\xi}{\L(\tau)}}
\approx \frac{\sum_{s=1}^N \L(\tau^s) \nabla_{\thetab^i} \log  \pib_{\thetab^i}(\ub_t^s)}{\sum_{s=1}^N \L(\tau^s)}
\end{align}

For a single particle ($m=1$), the full gradient in \eqref{eq:svgd_mpc}  reduces to
: $\bm{\phi}^*(\thetab) =  \nabla_\thetab \log p_t(\thetab|\O; \xi,\x_t) $. SVGD then produces a local MAP estimate of the posterior distribution over $\thetab$. As a result, the SV-MPC update step exhibits strong similarity to common MPC algorithms (such as MPPI~\cite{infotheoreticmpc} and CEM~\cite{cem}) depending on the chosen likelihood function, exhibiting equivalence under  parameter values and choice of prior. Discussion of these comparisons can be found in appendix~\ref{app:llh_examples}.

A significant advantage of the SV-MPC formulation is the robustness to highly-peaked posterior distributions. If particles are initialized poorly, or if the target posterior changes significantly between time-steps, many particles may find themselves in regions of low-probability. Indeed, this may occur frequently for likelihoods $p(\O | \thetab^i; \xi, \x_t)$ with exponentiated cost (see appendix~\ref{app:exp_util}, for example). However, the shared gradient terms in \eqref{eq:svgd_mpc} allow these particles to overcome this degeneracy quickly, while avoiding collapse due to the repulsive term (the reader may refer to Figure~1. in \cite{liustein} for an intuitive illustration of this phenomenon.) As a consequence, SV-MPC avoids the problem of particle depletion often encountered in Sequential Monte Carlo methods~\cite{kantas2009smc}. 

\textbf{Kernels for trajectories \quad}\label{sec:kernel}High-dimensional inference problems pose significant challenges for SVGD as the repulsive force given by the derivative of the kernel with respect to the inputs diminishes as the dimensionality increases~\cite{zhuo2018message}. Inspired by probabilistic graphical models and the conditional independence assumptions encoded in Markov random fields, we tackle this issue by devising a kernel that factorizes a high-dimensional input into a sum of kernels defined over cliques of dimensions. This allows the exploitation of the Markov structure of the trajectories to address the curse of dimensionality. For example, assume that the posterior over the parameters $\thetab$ satisfies the conditional independence relations encoded in a graph ${\cal G}=({\cal V},{\cal E})$, with vertices ${\cal V}$, and edges ${\cal E}$ such that $p(\thetab) \propto \prod_{d\in {\cal E}}\psi_d(\theta_d)\prod_{(dt)\in {\cal E}}\psi_{dt}(\theta_d, \theta_t)$, where $\psi_d(\theta_d)$ and $\psi_{dt}(\theta_d,\theta_t)$ are unary and pairwise potential functions respectively. We define the kernel over parameters as,
\begin{align}
k(\thetab, \thetab') &= \sum_{d\in{\cal V}}k(\theta_d, \theta_d)
+ \sum_{(d,t)\in {\cal E}}k(\theta_{(d,t)}, \theta'_{(d,t)})
\label{eq:kernel}
\end{align}
The kernel is a sum of positive semi-definite kernels, so the result is a valid reproducing kernel Hilbert space~\cite{Smola01} but less sensitive to the curse of dimensionality. In this paper we adopt the smooth RBF kernel; $k(\thetab, \thetab')=\exp\left\{-\|\thetab - \thetab'\|^2_2/h\right\}$, where $h$ is evaluated using the median heuristic on the set of particles: $h = \mathrm{med}\left(\{\thetab^i\}\right)^2 / \log m$. 

\begin{wrapfigure}{r}{.50\linewidth} \vspace{-25pt}
    \setlength\intextsep{0pt}
    \captionsetup{belowskip=0pt}
    \setlength{\belowcaptionskip}{-20pt}
    \begin{center}
	\includegraphics[width=.5\columnwidth]{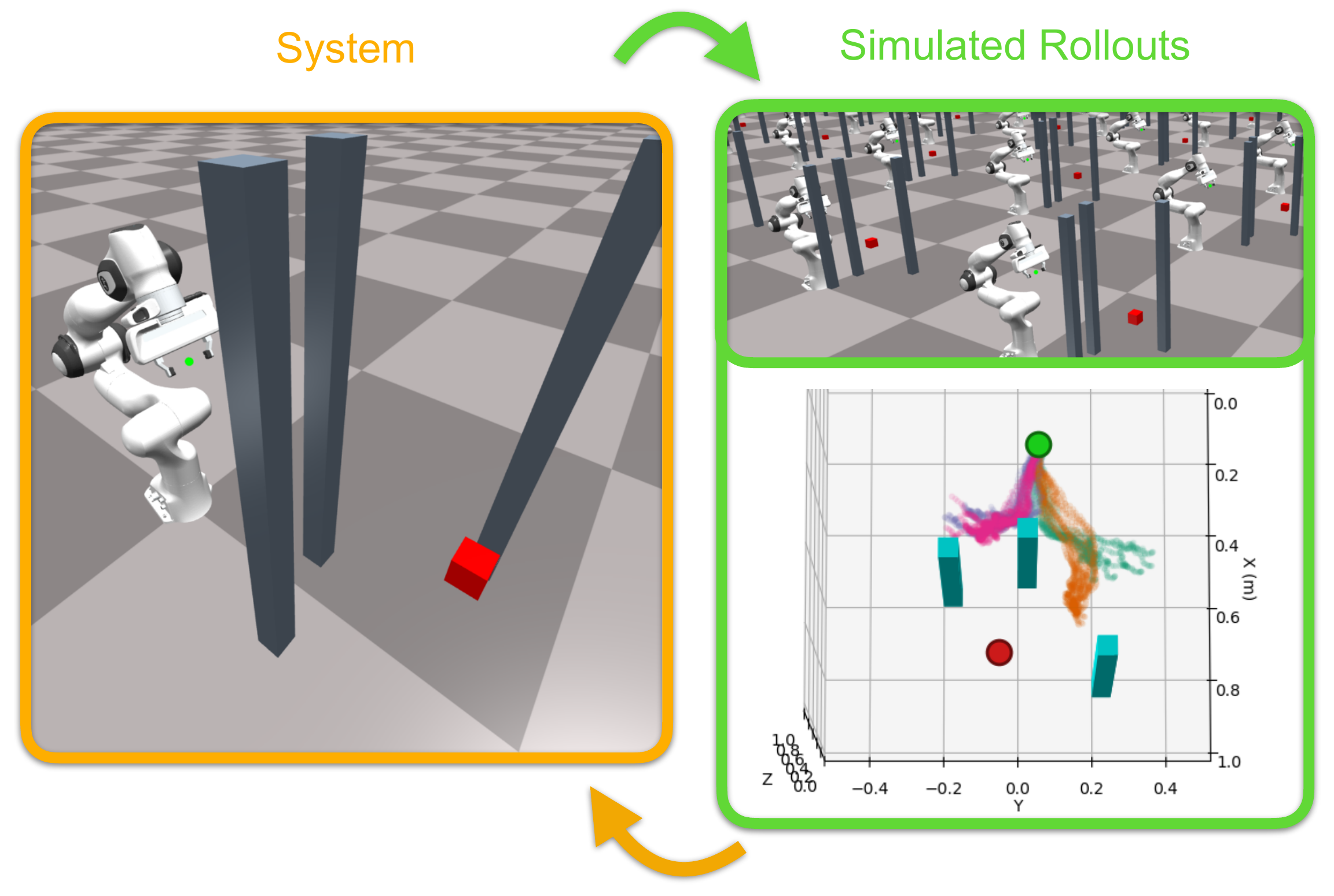}
	\captionsetup{size=footnotesize}
	\vspace{-10pt}
	\caption{\textbf{7-DOF manipulation task.} The SV-MPC framework is capable of reasoning over multi-modal distributions of trajectories in high-dimensional spaces. Here, the controller iteratively explores the posterior over joint-velocities by simulating trajectories in parallel (green frame) in order to guide the system (orange frame). Each particle-generated distribution is shown by a unique coloring over the generated state trajectories, as seen from a top-down view of the workspace. The robot arm manages to reach the goal (red), while avoiding poor local minima.}
	\vspace{-30pt}
	\label{fig:franka_reach}
	\end{center}
\end{wrapfigure}

\textbf{Action Selection \quad}\label{sec:action_selection}The variational inference procedure results in an approximation of the posterior distribution over $\thetab$. Following this step, a decision must be made on $\thetab_t$, such that a control action can be generated from the resulting policy $\pib_{\thetab_t}$ and executed on the real system. Here we outline two possible methods for choosing an appropriate $\thetab_t$.

We can first consider the relative probabilistic weight of the particles as an approximation to their posterior probabilities:
\begin{align}\label{eqn:best_action}
    w_i &= \frac{p(\O | \thetab^i; \xi, \x_t)\ \Tilde{q}_{t-1}(\thetab^i)} {\sum_{j=1}^m p(\O| \thetab^j; \xi,  \x_t)\ \Tilde{q}_{t-1}(\thetab^j)} 
    \\
    &\approx p(\thetab^i | \O ; \xi, \x_t).
\end{align}

One strategy to selecting $\thetab_t$ is to pick the highest-weighted particle $\thetab_t = \thetab_{i^*}$, which corresponds to the approximate MAP solution:
\begin{align}\label{eqn:sample_action}
    i^* &= \argmax_i [w_i] \\
    &\approx \argmax_i p(\thetab^i | \O ; \xi, \x_t).
\end{align}
An alternative approach is to randomly sample from the posterior distribution, which can be approximated by sampling from the set of particles according to their weight $w_i$. This can be performed by drawing from the categorical distribution over particle weights: $i^* \sim \mathrm{Cat}(w_i)$. 

\textbf{Shifting the distribution \quad}\label{sec:shifting}The prior $\Tilde{q}_t(\thetab; \x_t)$ is defined as the marginalized transition over the approximate posterior distribution obtained from the previous iteration:  $\Tilde{q}_t(\thetab; \x_t) =  \int p_t(\thetab| \thetab_{t-1} ; \x_t) q_{t-1}(\thetab_{t-1}) d\thetab_{t-1}$, with the transition probability $p_t(\thetab| \thetab_{t-1} ; \x_t)$ including a dependence on the currently observed state $\x_t$  for generality. This operation is akin to the prediction step commonly found in Bayes filtering and sequential Monte-Carlo methods, and can be interpreted as a probabilistic version of the shift operator defined in \cite{dmd-mpc} which serves to bootstrap the previous MPC solution to initialize the current iteration. Further details can be found in appendix~\ref{app:shift}.

 \textbf{Non-parametric SV-MPC \quad} In the setting where inference must be performed over open-loop controls $\thetab_t = \ub_t$, we can define the cost-likelihood as a function of state: $\L = \L(\xb_t)$. A cost on control can be replaced by defining an additional prior factor $p(\thetab)$, which can be combined with the transition probability above. For further details, refer to appendix~\ref{app:nonparam_mpc}. 
 
 \textbf{Trajectory optimization \quad}The variational inference framework defined by SV-MPC can be modified to accommodate general motion planning problems common to many robotics applications. This special case can be considered by (1) using the non-parametric formulation, (2) assuming a stationary posterior distribution (no shifting), and (3) defining a deterministic dynamics model $\bar{f}$ such that state trajectory probabilities can be represented as $p(\xb|\thetab)=\delta(\xb - \bar{f}(\thetab, x_0))$. A prior over sequences $p(\thetab)$ can be defined to encourage desired behavior such as smoothness~\cite{mukadam2018continuous}. Because individual particles are not guaranteed to produce a MAP estimate once the SVGD optimization has converged, they can be subsequently refined by applying stochastic gradient descent (SGD) using posterior gradients without kernelization. Selection of a feasible and optimal plan can be generated by simply picking the best particle. The resulting algorithm, SV-TrajOpt, is described in Algorithm~\ref{algo:svtrajopt}, with the likelihood gradient derived in appendix~\ref{app:motion_planning}.\vspace{-5pt}

\section{Experiments}\label{sec:experiments}\vspace{-5pt}
 In the following section, apply SV-MPC to common robotics problems: navigation, manipulation and locomotion. We end with an example of the SV-TrajOpt algorithm applied to a planar motion-planning scenario. All control algorithms were implemented in PyTorch, with batched gradient computation across particles and parallel generation of rollouts using either simulated or analytic dynamical models. Additional experimental details and results can be found in the appendix.
 
\begin{figure}[t]
	\centering
	\includegraphics[width=0.8\linewidth]{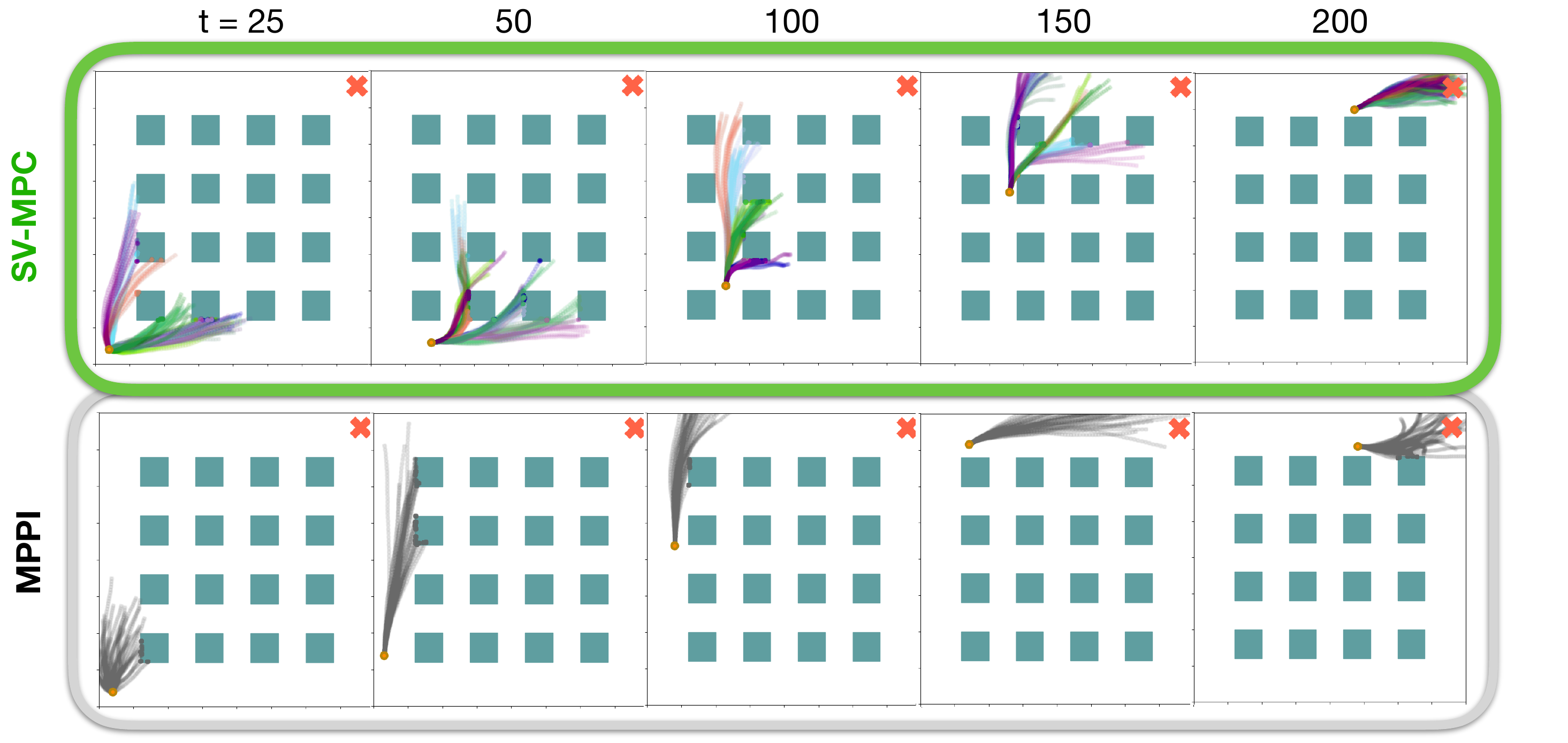}
	\captionsetup{size=footnotesize}
	\caption{\textbf{Planar Navigation Task.} The robot (orange dot) attempts to reach the goal location (red cross) while avoiding obstacles. Each frame depicts the environment state at a particular time-step, along with the distributions of sampled state-trajectory rollouts generated by the MPC controllers using the modeled dynamics. Each trajectory color is associated with a single particle from SV-MPC. The multi-modal distribution of SV-MPC is able to explore passages between obstacles and find shorter paths to the goal.}
	\label{fig:stein_vs_mpc}
	\vspace{-10pt}
\end{figure}

\textbf{Planar Navigation \quad}\label{sec:exp_planar_nav}  We construct a 2D robot navigation task, where a holonomic point-robot must reach a target location while avoiding obstacles (Figure~\ref{fig:stein_vs_mpc}). Hitting an obstacle will cause the agent to "crash" and prevent any further movement. This added non-differentiability makes the problem particularly challenging, but also suitable for sampling-based control schemes. The system exhibits stochastic dynamics, and is defined as a double-integrator model with additive Gaussian noise. The SV-MPC controller with exponentiated-utility (EU) is compared against MPPI (Figure~\ref{fig:stein_vs_mpc}) and CEM. We provide additional quantitative results in (appendix~\ref{app:additional_results}).

\begin{figure}[b]
    \centering
    \captionsetup{size=footnotesize}
    \includegraphics[width=0.9\linewidth]{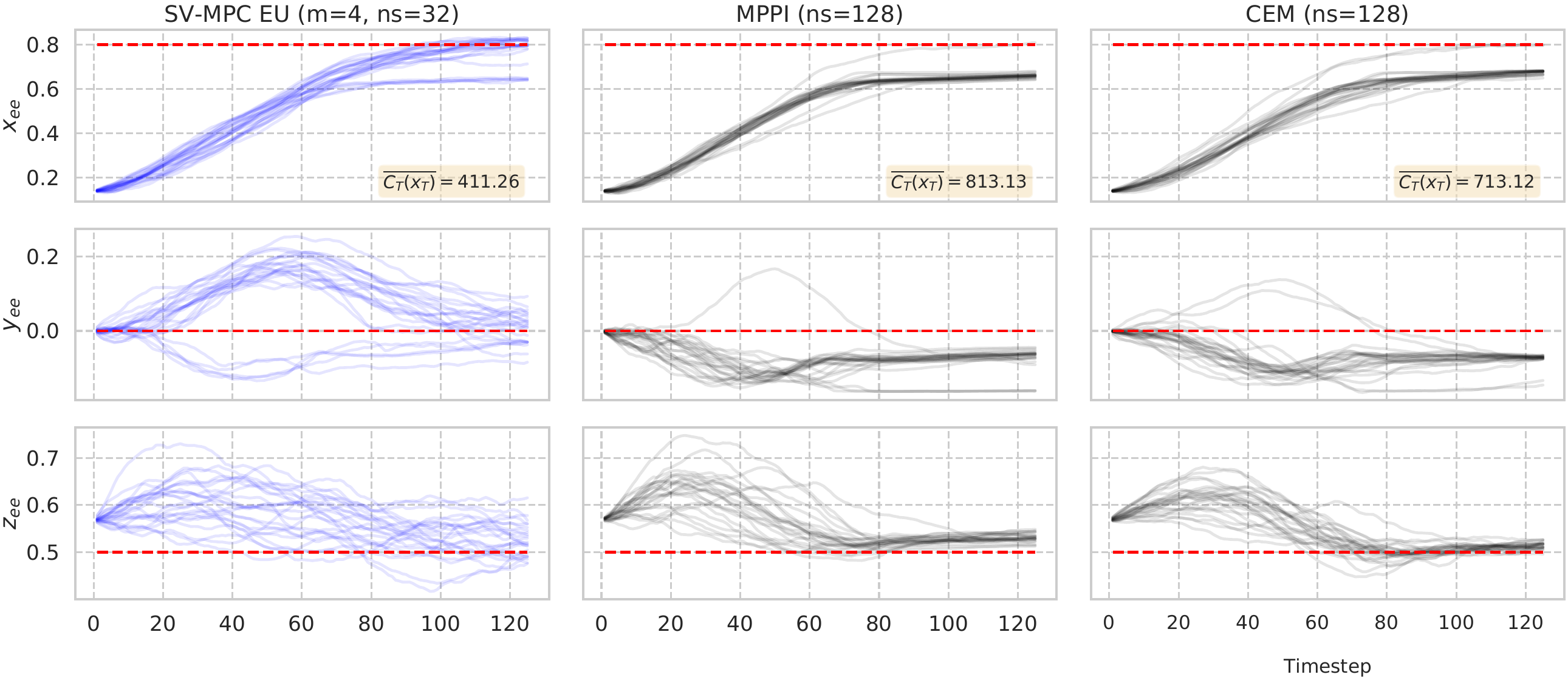}\vspace{-2mm}
    \caption{\textbf{Manipulation.} Examples of end-effector Cartesian trajectories resulting from application of different MPC algorithms on the Franka reaching experiment. The relative distance to the fixed target location is plotted over the length of each episode. The dashed red line indicates the coordinates of the target. The sample-averaged terminal cost for the final state $\overline{C_T(\x_T)}$ is evaluated over the 24 independent trials. With four particles ($m=4$), SV-MPC with Exponentiated utility likelihood manages to avoid bad local minima, despite higher-variance gradients due to fewer samples used to evaluate gradients ($\mathrm{ns}=32$ vs. $\mathrm{ns}=128$).}
    \label{fig:franka_results}
\end{figure}

\textbf{Manipulation \quad} We demonstrate SV-MPC on a 7-DOF reaching task (Figure~\ref{fig:franka_reach}). Velocity-based control commands are generated in the configuration space of a simulated robot manipulator, which must reach a stationary goal in its work-space. We leverage the GPU-accelerated Isaac-Gym library~\cite{isaac-gym} for parallel computation of trajectory rollouts in simulation during MPC iterations.

A sampling-based SV-MPC controller is compared against MPPI and CEM for an open-loop, constant-covariance Gaussian control distribution: $\pib_{\thetab} = \mathcal{N}(\ub_t; \thetab, \Sigma)$. The obstacles and the target are placed in order to demonstrate a local-minima trap: given a finite $H$, the optimal control solution is to move left. However, the opening between the obstacles in this direction is too narrow for the robot to move through. To avoid getting stuck, a sampling-based control scheme must generate a sequence which will move the robot in the other direction. For a uni-modal distribution, this may have a very low-probability, and recovery will not occur. Increasing the sampling covariance would mitigate this, but would require a larger amount of samples to reduce variance. 

Examples of trajectory executions are shown in Figure~\ref{fig:franka_results}. The total number of generated control samples is held constant across algorithms. Both MPPI and CEM tend towards the sub-optimal local minimum, leading the robot to get stuck between two obstacles. Using a particle-based representation of the posterior, SV-MPC can resolve multiple optima simultaneously, switching to lower-cost modes around obstacles and successfully reaching the goal.

\begin{figure}[t]
    	\centering
    	\captionsetup[subfigure]{size=footnotesize, justification=centering}
        \begin{subfigure}[b]{0.49\linewidth}
            \centering
        	\includegraphics[width=\linewidth]{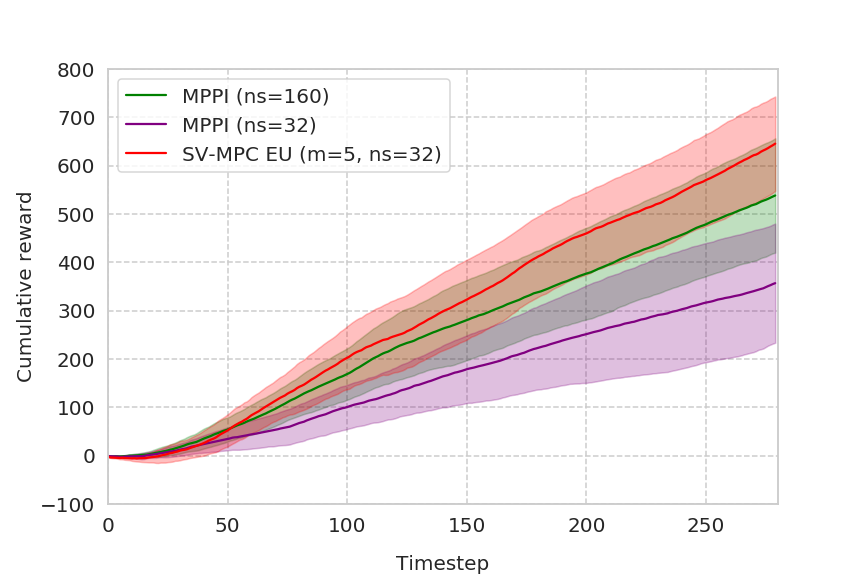}
        	\captionsetup{size=footnotesize}
        	\caption{\small Exponentiated Utility (EU)}
        	\label{subfig:halfcheetah_mppi}
    	\end{subfigure}
        \begin{subfigure}[b]{0.49\linewidth}
            \centering
        	\includegraphics[width=\linewidth]{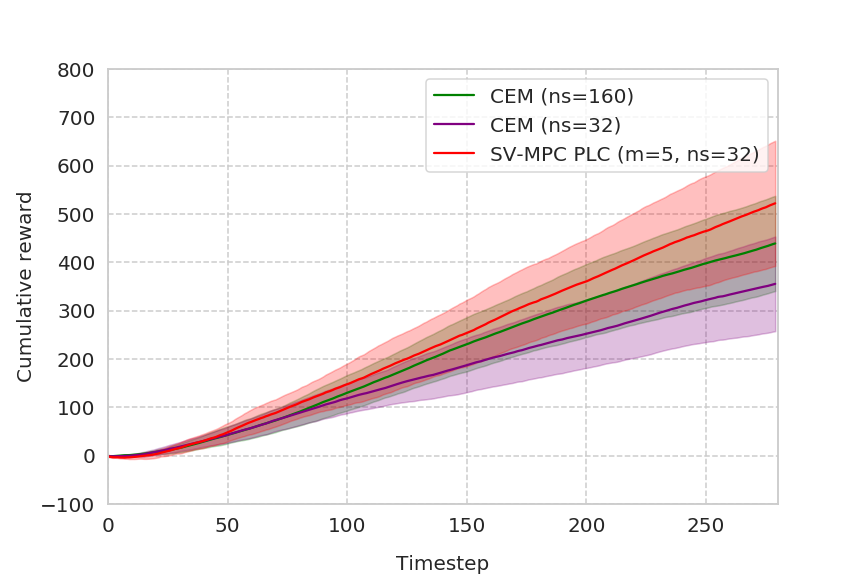}
        	\captionsetup{size=footnotesize}
        	\caption{\small Probability of Low-Cost (PLC)}
        	\label{subfig:cartpole}
    	\end{subfigure}
    	\captionsetup{size=footnotesize}
    	\caption{\textbf{Stochastic HalfCheetah.} Comparisons of cumulative-reward distributions over 16 independent trials, with mean and standard deviations shown. SV-MPC is capable of finding high-reward trajectories, using the same total amount of samples as MPPI and CEM.}
    	\label{fig:halfcheetah}\vspace{-5pt}
    \end{figure}

\textbf{Stochastic Half-Cheetah \quad}\label{app:exp_mpc_halfcheetah}To test our approach on a complex nonlinear system with discontinuous dynamics, we consider an environment common in many Reinforcement Learning benchmarks: the Half-Cheetah~\cite{brockman2016openai}. We use a stochastic version of the dynamics with additive noise in the control space. The cumulative rewards over multiple trials are plotted in Figure~\ref{fig:halfcheetah}.
  
\begin{figure}[t]
    \centering
    \captionsetup{size=footnotesize}
    \includegraphics[width=\linewidth]{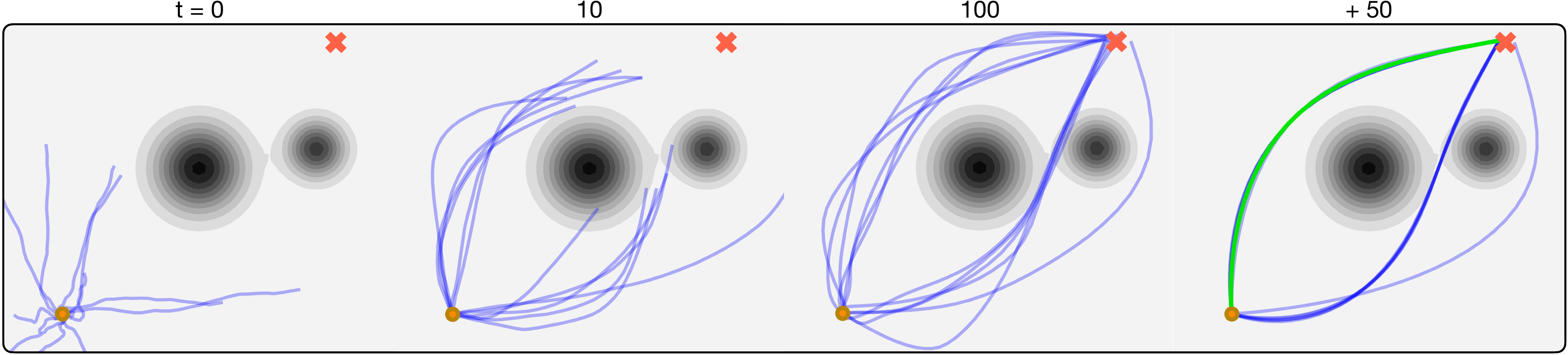}
    \caption{\textbf{Motion Planning.} SV-TrajOpt is applied to a velocity-controlled planar robot with fully-differentiable costs. Each blue state-trajectory results from a single particle control-sequence. Particles are randomly initialized from the prior ($t=0$), and are optimized until convergence ($t=10,\, 100$). Independent local MAP approximations are generated after 50 SGD iterations, with the lowest-cost particle shown in green.}
    \label{fig:planning_results}
    \vspace{-15pt}
\end{figure}

\textbf{Motion Planning \quad}\label{sec:exp_motion_planning} 
  We use the SV-TrajOpt algorithm to infer a distribution over optimal control sequences for a planar motion planning problem on a point robot (Figure~\ref{fig:planning_results}). The robot must find a velocity-based control sequence which results in a low-cost, feasible path around a set of obstacles to reach the goal. The occupancy map is fully-differentiable, allowing a gradient on obstacle cost to be computed numerically. A `smoothness' prior is defined over velocities as a multi-variate Gaussian with a tri-diagonal precision matrix, which favors low-acceleration trajectories.

\vspace{-10pt}
\section{Conclusion}\vspace{-5pt}
\label{sec:conclusion}

A novel formulation for Bayesian model predictive control is presented, where inference is performed directly over the control parameters and inputs. An algorithm for approximate inference is then proposed, where the posterior is represented as a set of particles and is optimized via SVGD. In contrast to pure Monte-Carlo sampling methods, gradient-based information can be exploited to improve particle efficiency, where many computationally-intensive operations can be run in parallel using effective GPU implementation. The flexibility of the approach can accommodate different cost transformations to modulate risk-seeking behavior, and can naturally be extended to trajectory optimization problems. We compare against common MPC baselines, demonstrating improved performance on a variety of control tasks. 
 
\bibliography{references}

\onecolumn
\begin{appendices}

\section{Notation}

The table below summarizes notation used throughout the paper.

\begin{table}[h]
  \centering
    \caption{Notations and definitions}
    \begin{tabular}{c|l}
    \toprule
       Notation & \multicolumn{1}{|l}{Description} \\
      \hline    
      \rule{0pt}{3ex}
      $\x_t$ & System state at time $t$ \\
      $\xb_t$ & State trajectory: sequence of length $H+1$, from time $t$\\
      $\u_t$ & Control input at time $t$ \\
      $\ub_t$ & Control trajectory: sequence of length $H$, from time $t$\\
      $\tau$ & State-control trajectory tuple, $\tau = (\xb_t, \ub_t)$\\
      $\hat{f}_\xi(\cdot, \cdot)$ & Modeled system dynamics, parameterized by $\xi$\\
      $C(\cdot, \cdot)$ & Total cost for horizon $H$ \\
      $\pi_{\theta_t}(\cdot)$ & Policy at time $t$ parameterized by $\theta_t$\\
      $\thetab_t$ & Sequence of policy parameters \textit{or} control inputs \\
      $\pib_{\thetab_t}(\cdot)$ & Sequence of control policies \\
      $\L(\cdot)$ & Cost-likelihood function\\
      $q(\cdot)$ & Approximate posterior distribution\\ 
      $\phi(\cdot)$ & Perturbation direction or velocity field that decreases \\
      & the KL divergence between the particle distribution \\
      & and target distribution\\
      $\hat{\phi}(\cdot)$ & Approximated perturbation direction over particles\\ 
      $k(\cdot, \cdot)$ & Kernel specifying a reproducing kernel Hilbert space \\
    \bottomrule
    \end{tabular}
  \label{table:notation}
\end{table}

\section{Algorithm}\label{app:algorithm}

\definecolor{alg}{RGB}{255, 137, 137}

{\SetAlgoNoLine%
  \begin{algorithm}
    \DontPrintSemicolon 
    {\color{alg} * Components which apply only in the presence of a parameterized policy $\pib$ are marked in red.}\\[3pt]
    \KwIn{
        Initial state $\x_0$, 
        dynamics $\hat{f}_\xi$,
        cost-likelihood $\L$,
        prior $p_0(\thetab; \cdot)$,
        kernel $k$,
        {\color{alg}policy $\pib$}
        }
    \vspace{10pt}
    Initialize $\Tilde{q}_0(\thetab)= p_0(\thetab; \x_0)$\\[3pt]
    Sample $\{\thetab^i\}_{i=1}^m \sim \Tilde{q}_0(\thetab) $\\[3pt]
    \For{$t = 0, 1, ...,T-1$ } {
    \ForPar{$i = 1, 2, ...,m$}{
    $\nabla_{\thetab^i}\log p_t( \thetab^i |  \O;  \xi, \x_t) = \nabla_{\thetab^i} \log \expect{{\color{alg}\pib_{\thetab^i},} \hat{f}_\xi}{\L(\tau)} + \nabla_{\thetab^i} \log \Tilde{q}_t(\thetab^i)$
    }
    \ForPar{$i = 1, 2, ...,m$ } {
    $\Delta \thetab^i \leftarrow \frac{1}{m}\sum_{j=1}^m k(\thetab^j, \thetab^i)\nabla_{\thetab^j}\log p_t( \thetab^j |  \O;  \xi, \x_t) + \nabla_{\thetab^j} k(\thetab^j, \thetab^i)$ \\[5pt] 
    $\thetab^i \leftarrow \thetab^i + \epsilon \Delta \thetab^i$}
    $w^i \leftarrow p(\O | \thetab^i; \x_t) \Tilde{q}_t(\thetab^i)$\\[5pt]
    $w^i \leftarrow \frac{w^i}{\sum_{j=1}^m w^j}$ \\[5pt]
    Pick $\thetab^*$ (using \eqref{eqn:best_action} or \eqref{eqn:sample_action}, for example)\\
    Get control input: $\u_t \leftarrow \thetab^*$ \ {\color{alg} or sample $\u_t\sim \pib_{\thetab^*_t}(\x_t)$}\\
    Sample true dynamics: $\quad \x_{t+1} \sim f (\x_t, \u_t) $ \\
    Shift particles: $\quad \Tilde{\thetab}^i = \Phi(\thetab^i)$\\
    Update prior: $\quad \Tilde{q}_{t+1}(\thetab; \x_{t+1}) = \sum_{i=1}^m w^i p_t(\thetab| \Tilde{\thetab}^i ; \x_{t+1})$
    }
    \caption{Stein Variational MPC}
    \label{algo:svmpc}
  \end{algorithm}}%
  
  {\SetAlgoNoLine%
  \begin{algorithm}
    \DontPrintSemicolon 
    \KwIn{
        Initial state $\x_0$, 
        dynamics $f$,
        cost function $\cost$,
        prior $p_0(\thetab)$,
        kernel $k$,
        termination condition $\mathrm{Done}(\cdot)$
        }
    \vspace{10pt}
    Sample $\{\thetab^i\}_{i=1}^m \sim p_0(\thetab) $\\[5pt]
    Set $\Delta \thetab^i = \mathrm{Inf}\ \forall\  i \in 1:m$\\[5pt]
    \While{$\mathrm{Done}(\{\Delta\thetab^i\}) \textrm{ is False}$} {
    \ForPar{$i = 1, 2, ...,m$}{
    Forward dynamics : $\xb^i = \xb(f, \thetab^i, \x_0)$\\[5pt]
    $\nabla_{\thetab^i}\log p( \thetab^i |  \O;  \xi) = - \alpha \nabla_{\thetab^i} \cost(\xb^i) + \nabla_{\thetab^i} \log p_0(\thetab^i)$
    }
    \ForPar{$i = 1, 2, ...,m$ } {
    $\Delta \thetab^i \leftarrow \frac{1}{m}\sum_{j=1}^m k(\thetab^j, \thetab^i)\nabla_{\thetab^j}\log p( \thetab^j |  \O;  \xi) + \nabla_{\thetab^j} k(\thetab^j, \thetab^i)$ \\[5pt] 
    $\thetab^i \leftarrow \thetab^i + \epsilon \Delta \thetab^i$}
    }\vspace{5pt}
    (Optional: SGD refinement) \\
    \For{$iter = 1,..., N$ } {
    $\Delta \thetab^i \leftarrow \nabla_{\thetab^i}\log p( \thetab^i |  \O;  \xi) \quad \forall\  i \in 1:m$\\[5pt] 
    $\thetab^i \leftarrow \epsilon_r \Delta \thetab^i \quad \forall\  i \in 1:m $
    }\vspace{10pt}
    $\thetab^* = \argmax_{\thetab_i} \log p(\thetab^i | \O ; \xi)$ \\[5pt]
    \caption{SV-TrajOpt}
    \label{algo:svtrajopt}
  \end{algorithm}}%
  
\section{Shift operation}\label{app:shift}
As described in Section \ref{sec:shifting}, the proposal distribution $q$ is propagated after each round of MPC by marginalizing over the previous solution:
\begin{align}
    \Tilde{q}_{t+1}(\thetab; \x_{t+1}) =  \int p_t(\thetab| \thetab_t ; \x_{t+1}) q_t (\thetab_t) d\thetab_t
\end{align}
given a transition probability, $p_t(\thetab| \thetab_{t-1} ; \x_t)$. This operation serves to approximate the prior at the new iteration. Given the empirical distribution $q_t$, we can simplify the expression above:
\begin{align}\label{eq:prior_update}
    \Tilde{q}_{t+1}(\thetab; \x_{t+1}) &=  \sum_{i=1}^m \int p_t(\thetab| \thetab_t ; \x_{t+1})\ w^i\delta_{\thetab_i}(\thetab_t) d\thetab_t = \sum_{i=1}^m w^i p_t(\thetab| \thetab^i ; \x_{t+1})
\end{align}
resulting in a mixture of conditional probabilities.

In many open-loop MPC implementations~\cite{infotheoreticmpc, dmd-mpc}, it is assumed that the solution does not change significantly between each round, given an accurate dynamics model and sufficiently high controller frequency to resolve the dynamics and possible perturbations. This motivates a common heuristic used to reduce the computational burden between timesteps, which is to shift the control distribution $\pib_{\thetab_t}$, forward-in-time by one step. That is, given an initial parameter sequence $\thetab_t=\left(\theta_t, \theta_{t+1}, ..., \theta_{t+H-1} \right)$:
\begin{align}
    \Tilde{\thetab}_{t+1} &= \Phi(\thetab_t) \\
    &= (\theta_{t+1}, \theta_{t+2}, ..., \theta_{t+H-1}, \Tilde{\theta}_{t+H-1} )
\end{align}
where the new parameter $\Tilde{\theta}_{t+H-1}$ is chosen to reflect the expected final action. 
In the implementation, we adopt a similar heuristic, where the empirical distribution is first shifted deterministically according to:
\begin{align}
    \Tilde{\thetab}^i &= \Phi(\thetab^i) \quad \forall \quad i \in 1:m
\end{align}
and setting the resulting distribution as $\Tilde{q}_t(\thetab) = q_t(\thetab)$. The shifted particles, $\Tilde{\thetab}^i$, are then used in the following iteration to approximate the posterior, and the prior is updated according to \eqref{eq:prior_update}. The shift operation can then be summarized by the following two sub-steps:
\begin{enumerate}
\item Shift particles: $\quad \Tilde{\thetab}^i = \Phi(\thetab^i)$
\item Update prior: $\quad \Tilde{q}_{t+1}(\thetab; \x_{t+1}) = \sum_{i=1}^m w^i p_t(\thetab| \Tilde{\thetab}^i ; \x_{t+1})$
\end{enumerate}

\section{Variational Inference}\label{app:vi_objective}

In variational inference, a target distribution $  p(\thetab | \O) $ is approximated by a candidate  $q^*(\thetab)$, belonging to a specified family of distributions $\mathcal{Q} = \{ q(\thetab) \}$, by minimizing the KL-divergence:
\begin{align}
q^* &= \argmin_{q\in \mathcal{Q}}\ D_{KL} \left( q(\thetab) ||  p(\thetab | \O) \right) 
\end{align}
The optimal distribution $q^*$ is also the solution to the following objective:
\begin{align}
q^* = \argmax_{q\in \mathcal{Q}}\ \expect{q}{\log p(\O | \thetab)}  - D_{KL} \left( q(\thetab)\ ||\ p(\thetab)\right)
\end{align}
\textit{Proof 1}:
\begin{align}
q^* &= \argmin_{q\in \mathcal{Q}}\ D_{KL} \left( q(\thetab) ||  p(\thetab | \O) \right) \\
&= \argmin_{q\in \mathcal{Q}}  \int \log q(\thetab)  dq(\thetab)  - \int \log p(\thetab | \O)  dq(\thetab) \\
&= \argmin_{q\in \mathcal{Q}}  \int \log q(\thetab)  dq(\thetab)  - \int \log p(\O | \thetab)  dq(\thetab)  -  \int \log p(\thetab)  dq(\thetab)  + \log Z\\
&= \argmin_{q\in \mathcal{Q}}  \int \log q(\thetab)  dq(\thetab)  - \int \log p(\O | \thetab)  dq(\thetab)  -  \int \log p(\thetab)  dq(\thetab) \\
&= \argmin_{q\in \mathcal{Q}}\ \expect{q}{\log q(\thetab)}  - \expect{q}{\log p(\O | \thetab) + \log p(\thetab)} \\
&= \argmax_{q\in \mathcal{Q}}\ \expect{q}{\log p(\O | \thetab)}  - D_{KL} \left( q(\thetab)\ ||\ p(\thetab)\right)
\end{align}

where $Z$ is the normalizing constant. The functional in the objective is also known as the variational free energy $F_q(\thetab) := \expect{q}{\log p(\O | \thetab)}  - D_{KL} \left( q(\thetab)\ ||\ p(\thetab)\right)$.

An alternative derivation, which is relevant to the discussion in Appendix~\ref{app:pi_connection}, can be found by deriving the lower-bound to the log marginal-likelihood:\\

\textit{Proof 2}:
\begin{align}\label{eq:elbo_proof_2}
\log \expect{p}{ p(\O | \thetab)} &= \log \int p(\O | \thetab) p(\thetab) d\thetab \\ 
&= \log \int p(\O | \thetab) p(\thetab) d\thetab \\
&= \log \int p(\O | \thetab) \frac{p(\thetab)}{q(\thetab)} q(\thetab) d\thetab \\
&= \log \expect{q}{p(\O | \thetab) \frac{p(\thetab)}{q(\thetab)} } \\
&\geq \expect{q}{\log \frac{p(\O | \thetab) p(\thetab)}{q(\thetab)}} \label{eq:jensen}\\
&= \expect{q}{\log p(\O | \thetab)} - \expect{q}{\log \frac{p(\thetab)}{q(\thetab)}} \\
&= \expect{q}{\log p(\O | \thetab)}  - D_{KL} \left( q(\thetab)\ ||\ p(\thetab)\right)
\end{align}
where \eqref{eq:jensen} results from log-concavity and application of Jensen's inequality. 
Equality is obtained when $q$ matches the posterior probability. In other words, provided that $p(\thetab | \O) \in \mathcal{Q}$, we have:
\begin{align}
q^*(\thetab) = \frac{p(\O | \thetab)\ p(\thetab)}{\int p(\O | \thetab)\ p(\thetab) d\thetab} = p(\thetab | \O)
\end{align}
As such, we can define the minimization:
\begin{align} \label{eq:vi_ineq}
- \log \expect{p}{ p(\O | \thetab)} &= \min_{q \in \mathcal{Q}} -\expect{q}{\log p(\O | \thetab)}  + D_{KL} \left( q(\thetab)\ ||\ p(\thetab)\right)
\end{align}

\section{Connection to Path Integral Control}\label{app:pi_connection}

The equation derived in \eqref{eq:vi_ineq} is well-known in statistical thermodynamics, where the random variable under consideration is the energy $C(x): \Omega_x \mapsto \mathbb{R}$, a non-negative real-valued measurable property, and $x\in \mathbb{R}^n$ is the state of the system. 
The Helmholtz free energy of $C$, with respect to probability density $p$, is defined as the function :
\begin{align}
    -\frac{1}{\alpha}\log \expect{p}{\exp \left(-\alpha C(x) \right)},
\end{align}
where $\alpha > 0$ is the (inverse) temperature. The corresponding variational inequality, known as the Donsker-Varadahn principle~\cite{varadhan1984large, hartmann2017variational}, relates the free energy as the Legendre transform of the entropy:
\begin{align}
-\frac{1}{\alpha}\log \expect{p}{\exp \left(-\alpha C(x) \right)} &= \min_q\ -\frac{1}{\alpha} \expect{q}{\log \exp \left(-\alpha C(x) \right)}  + \frac{1}{\alpha} D_{KL} \left( q(x)\ ||\ p(x)\right) \\
&= \min_q\ \expect{q}{C(x)}  +  \frac{1}{\alpha} D_{KL} \left( q(x)\ ||\ p(x)\right) \label{eq:pi_objective}
\end{align}
Following the same derivation presented in \eqref{eq:elbo_proof_2}-\eqref{eq:vi_ineq}, the solution is then found to be the Gibbs distribution:
\begin{align}
q^* = \frac{\exp \left(-\alpha C(x) \right) p(x)}{\int \exp \left(-\alpha C(x) \right) p(x)  dx}.
\end{align}
In the context of stochastic dynamics, the above is also applicable to random paths generated by a Markov diffusion process \cite{hartmann2017variational}. This can then be extended to optimal control, by addressing the KL-minimization problem between controlled and uncontrolled stochastic systems. The connection was developed and explored for continuous-time dynamics in previous work, such as \cite{theodorou2012duality, theodorou2015nonlinear}. Here, the nonlinear-affine dynamics under consideration are subject to Brownian motion:
\begin{align}
    \x_{t+1} &= \bar{f}(\x_t) + G(\x_t)\u_t + B(\x_t)\w_t, \quad \w_t \sim \mathcal{N}(0, \Sigma) \\
    &= f(\x_t, \u_t)
\end{align}
for the discrete-time case. For a SOC expected-cost objective:
\begin{align}
    \min_{U_t}J(\xb_t) = \min_{U_t=(\u_0,...,\u_{T})}\expect{f}{c_{\mathrm{term}}(\x_T) + \sum^T_{t=0} c_t(\x_t) + \frac{1}{2}\u_t^\top R\u_t }\ ,
\end{align}

the optimal control distribution could be derived by a change of measure using Girsanov's theorem, which was shown to satisfy the Hamilton-Jacobi-Bellman (HJB) equations for optimality. To solve the HJB equations, an exponentiated form of the value function, the desirability-function, was defined: $\Psi(\x_t) = \exp \left(-\alpha \mathrm{V}(\x_t) \right)$. By application of the Feynman-Kac lemma, the transformed HJB partial differential equation can be solved as:
\begin{align}
    \Psi(\x_t) &= \int \exp \left(-\alpha C(\xb_t) \right) p_f(\xb_t; \x_t)  d\xb_t \\
    &= \expect{p_f}{\exp \left(-\alpha C(\xb_t) \right)}
\end{align}

where $p_f$ denotes the passive dynamics \textit{i.e.} the probability density of trajectories $\xb_t$, resulting from the stochastic dynamics $f(\x_t,\u_t)$ with $\u_t=0$. Under the assumption that $R=\frac{1}{\alpha}\Sigma^{-1}$, this can then be used to evaluate the optimal control law:
\begin{align} \label{eq:pi_opt_ctrl}
 \u^*_t &= - \alpha \Sigma\, G(\x_t)^\top \nabla_\x \mathbf{V}(\x_t) \nonumber\\
 &=  \Sigma\,G(\x_t)^\top\nabla_\x \log \Psi(\x_t) \nonumber\\
 &=  \Sigma\, G(\x_t)^\top \frac{\nabla_\x \Psi}{\Psi} \ . 
\end{align}
For receding-horizon control, we can borrow from \cite{infotheoreticmpc} and consider a discrete-time case of Path-Integral control for a nonlinear stochastic dynamical system: 
\begin{align}
    \x_{t+1} = \Bar{f}(\x_t, \mathbf{v}_t), \ \mathbf{v}_t \sim p(\mathbf{v}_t | \u_t)
\end{align}
with nominal deterministic dynamics $\Bar{f}$,  commanded control input $\u_t$, and stochastic perturbations $\mathbf{v}_t$ which are exhibited in the control input channel. The uncontrolled system is then realized when $\u_t=0$, and is controlled otherwise. Given a sequence of perturbations: $V_t = \left(\mathbf{v}_t, \mathbf{v}_{t+1}, ... \mathbf{v}_{t + H-1} \right)$, we can write the resulting state trajectory as $\xb_t = F(\x_t, V_t)$, where $F$ performs consecutive application of the dynamics $\Bar{f}$ given $\x_t$ and a sequence $V_t$. We can then consider probability distributions directly over $V_t$, with $p(V_t)$ for uncontrolled dynamics and $q(V_t | U_t)$ for the controlled system, with the sequence of control inputs given by $U_t = \left(\u_t, \u_{t+1}, ... \u_{t + H-1} \right)$. A cost on state $C(X_t)$ is mapped to the random variable $V_t$ by the convolution $S = C \circ F$. Similarly to equations \eqref{eq:vi_ineq} and \eqref{eq:pi_objective} above, we then seek a solution to minimize the variational objective:
\begin{align}
q^* &= \argmin_{q}\ D_{KL} \left(q\,||\,q^*(V_t)\,  \right)\\
&=  \argmin_{q}\ - \frac{1}{\alpha}\expect{q}{\log \exp \left(-\alpha S(V_t)\right)}  +  \frac{1}{\alpha}D_{KL} \left( q\ ||\ p\right)\\
&= \argmin_{q}\ \expect{q}{S(V_t)}  + \frac{1}{\alpha} D_{KL} \left( q\ ||\ p\right) 
\end{align}
where the optimal distribution is then known to be
\begin{align}
    q^*(V_t) &= \frac{\exp \left(-\alpha S(V_t) \right) p(V_t) }{\int \exp \left(-\alpha S(V_t) \right) p(V_t) dV_t} \ .
\end{align}

Although we have the form of the optimal control distribution, an analytic solution is generally intractable due to the partition function. The algorithm for Model Predictive Path Integral Control (MPPI) \cite{williams2016aggressive, infotheoreticmpc} proceeds by defining a surrogate cross-entropy minimization problem:
\begin{align}
    \min_{q\in \mathcal{Q}}\ D_{KL} \left( q^*(V_t)\, ||\, q\right) 
\end{align}
for a tractable family of distributions $\mathcal{Q}$, typically fixed-covariance Gaussians with mean parameters $\mathbf{\mu}_t=U_t$ and an equivalent assumption on the controlled dynamics distribution. 

In the SV-MPC framework, the original variational objective of Path Integral control can be addressed, where
\begin{align}
    \min_{q \in \mathcal{Q}}\ D_{KL} \left(q\,||\,q^*(V_t)\,  \right). 
\end{align}
We can apply the non-parametric Bayesian-MPC formulation (Setion \ref{sec:nonparam_mpc}) to solve for $q^*(V_t)$ directly over $V_t$, setting the inference parameter to be $\thetab = V_t$ with a prior given by the passive dynamics $p(\thetab; \x_t) = p(V_t)$. The likelihood term $p(\O|\thetab; \xi,\x_t)$ can be derived by using the exponentiated-utility $\L(\xb_t) = \exp \left(-\alpha C(\xb_t)\right)$ and defining a dirac measure over state trajectories to denote the probability density of $\xb_t$ given $\thetab$:
\begin{align}
p(\O|\thetab; \xi,\x_t) &\propto \int \exp \left(-\alpha C(\xb_t)\right) \delta (\xb_t - F(\x_t, \thetab))d\xb_t \\
&= \exp \left(-\alpha C(F(\x_t, \thetab))\right)\\
&= \exp \left(-\alpha S(\thetab) \right) \ .
\end{align}
where we drop $\xi$-notation for the dynamics parameters for simplicity.  

We then recover the Bayesian formulation:
\begin{align}
q^*(\thetab) &= \frac{p(\O|\thetab; \xi,\x_t)\ p(\thetab; \x_t)}{\int p(\O|\thetab; \xi,\x_t)\ p(\thetab; \x_t) d\thetab} \\
&= \frac{\exp \left(-\alpha S(\thetab) \right) \ p(\thetab; \x_t)}{\int \exp \left(-\alpha S(\thetab) \right) \ p(\thetab; \x_t) d\thetab}\\
&= p(\thetab | \O; \xi,\x_t)
\end{align}
Approximate inference on the posterior $p(\thetab | \O; \xi,\x_t)$ can then be performed using  \hyperref[app:algorithm]{Algorithm 1}.

We can further consider the special case of control-affine dynamics of the form:
\begin{equation}
    \x_{t+1} = \bar{f}(\x_t) + G(\x_t)(\u_t + \mathbf{v}_t), \quad \mathbf{v}_t \sim \mathcal{N}(0, \Sigma) .\\
\end{equation}
with passive dynamics: 
\begin{equation}
\x_{t+1} = \bar{f}(\x_t) + G(\x_t) \mathbf{v}_t, \quad \u_t=0\ \forall \ t \ .
\end{equation}
We refer to this in trajectory-wise form as $\xb_t = F(\x_t, V_t)$. 

By application of \eqref{eq:pi_opt_ctrl}, the optimal control action can then be determined :
\begin{align}
 \u^*_t = \Sigma\, G(\x_t)^\top \frac{\nabla_\x \Psi}{\Psi} &= \Sigma\, G(\x_t)^\top \frac{\nabla_\x  \expect{V_t\sim p(V_t)}{\exp(-\alpha S(V_t))}}{\expect{V_t\sim p(V_t)}{\exp(-\alpha S(V_t))}}  \\
 &= - \alpha \Sigma\, G(\x_t)^\top \frac{\expect{V_t\sim p(V_t)}{\exp(-\alpha S(V_t)) \nabla_\x S(V_t) |_{\x=\x_t}}}{\expect{V_t\sim p(V_t)}{\exp(-\alpha S(V_t))}}  \\
 &\approx - \alpha  \Sigma\, G(\x_t)^\top \frac{\sum^m_{i=1} \exp(-\alpha S(\thetab_i)) \nabla_\x S(\thetab_i) |_{\x=\x_t}}{\sum^m_{i=1} \exp(-\alpha S(\thetab_i) )} 
\end{align}

where $\nabla_\x S(\thetab_i) |_{\x=\x_t}$ is the gradient of the path cost with respect to the current state $\x_t$. This can be expressed in terms of the nominal dynamics by using the chain rule:
\begin{equation}
    \nabla_\x S(\thetab_i)^\top |_{\x=\x_t} = \deriv{C(\xb_t)}{\xb_t} \bigg|_{\xb_t=F(\x_t, \thetab_i)} \cdot \deriv{F(\x, \thetab_i)}{\x} \bigg|_{\x=\x_t}
\end{equation}
Numerical evaluation of this gradient can be performed via backpropagation for each $i$-th particle.

\section{Likelihood functions}\label{app:llh_examples}

\textbf{Exponentiated Utility (EU) \quad}\label{app:exp_util} By picking the cost-likelihood to be: $\L(\tau) = \exp \left( -\alpha C(X_t, U_t)\right)$, we obtain the likelihood function 
\begin{align}
\expect{\pib_{\thetab}, \hat{f}_\xi}{\ \L(\tau)}= \expect{\pib_{\thetab}, \hat{f}_\xi}{\exp\left( -\alpha C(X_t, U_t)\right)}.
\end{align}
This is otherwise known as the Free-Energy of the cost function $C(\x_t, \u_t) $ \cite{theodorou2012duality}, as well as the "soft maximum" or "risk-aware" loss. This yields the likelihood-gradient:
\begin{align}
\nabla_{\thetab} \log \expect{\pib_{\thetab}, \hat{f}_\xi}{\ \L(\tau)} &= 
\nabla_\thetab  \log\expect{\pib_{\thetab}, \hat{f}_\xi}{\exp\left(-\alpha C(X_t, U_t)\right)}\\[3pt]
&= \frac{\expect{\pib_{\thetab}, \hat{f}_\xi}{\exp\left(-\alpha C(X_t, U_t)\right)\nabla_\thetab\log  \pib_{\thetab}}}{\expect{\pib_{\thetab}, \hat{f}_\xi}{\exp\left(-\alpha C(X_t, U_t)\right)}}
\end{align}

As $\alpha \rightarrow \infty$, regions of high-cost are assigned lower probability, making the distribution of resulting policies  risk-averse. Conversely, as $\alpha \rightarrow 0$,  high-cost regions have higher likelihood, making the policy distributions more risk-seeking.

With this form of the gradient, we can choose a control policy as a sequence of Gaussian distributions over open-loop controls with fixed covariance: given that $\pib_{\thetab_t} =  \left( \pi_{\theta_t}, \pi_{\theta_{t+1}}, ..., \pi_{\theta_{t+H-1}}\right)$, we have the instantaneous policy $\pi_{\theta_t} = \mathcal{N}(\mu_t, \Sigma_t)$, our decision parameter is thus $\thetab_t=\bm{\mu}_t = \left(\mu_t, \mu_{t+1}, ... \mu_{t+H-1}\right)$, where $\mu_t \in \mathbb{R}^d$, $\Sigma_t \in \mathbb{R}^{d\times d}$.  Considering the SV-MPC step in \eqref{eq:svgd_mpc} for a single particle, we can derive the update for a parameter element  $h \in (t, t+1, ..., t+H-1)$:
\begin{align}
\theta'_h &= \theta_h + \epsilon \Big(\nabla_{\theta_h} \log p_t(\O\,|\,\thetab; \xi, \x_t) + \nabla_{\theta_h} \log \Tilde{q}_t(\thetab) \Big) \\
&= \theta_h + \epsilon\,\frac{\expect{\pib_{\thetab}, \hat{f}_\xi}{\exp\left(-\alpha C(X_t, U_t)\right)\nabla_{\theta_h}\log  \pi_{\theta_h}}}{\expect{\pib_{\thetab}, \hat{f}_\xi}{\exp\left(-\alpha C(X_t, U_t)\right)}} + \epsilon \nabla_{\theta_h} \log \Tilde{q}_t(\thetab) \\
&= \theta_h + \epsilon\,\Sigma^{-1}_t \frac{\expect{\pib_{\thetab}, \hat{f}_\xi}{\exp\left(-\alpha C(X_t, U_t)\right)(\u_h - \theta_h)}}{\expect{\pib_{\thetab}, \hat{f}_\xi}{\exp\left(-\alpha C(X_t, U_t)\right)}} + \epsilon \nabla_{\theta_h} \log \Tilde{q}_t(\thetab)\\
&= (I - \epsilon\,\Sigma^{-1}_t )\,\theta_h + \epsilon\,\Sigma^{-1}_t \frac{\expect{\pib_{\thetab}, \hat{f}_\xi}{\exp\left(-\alpha C(X_t, U_t)\right)\u_h}}{\expect{\pib_{\thetab}, \hat{f}_\xi}{\exp\left(-\alpha C(X_t, U_t)\right)}} + \epsilon \nabla_{\theta_h} \log \Tilde{q}_t(\thetab)  
\end{align}
where the gradient of the log-prior will depend on the particular choice of transition probability, $p_t(\thetab| \Tilde{\thetab} ; \x_{t+1})$, used in the shift operation (see Appendix~\ref{app:shift}). The update reduces to that found in MPPI~\cite{infotheoreticmpc} if we consider the control distribution to be  uncorrelated across dimensions (as is often done in practice): $\Sigma_t = \sigma_t^2 I$. Setting the step-size to $\epsilon = \sigma^2$, and assuming a uniform prior on controls, $p_t(\theta| \Tilde{\theta} ; \x_{t+1}) = \mathcal{U}(\theta_{min}, \theta_{max})$:
\begin{align}
\theta'_h &= (I - \sigma_t^2\,\sigma^{-2}_t I)\,\theta_h + \sigma_t^2\,\sigma^{-2}_t I \frac{\expect{\pib_{\thetab}, \hat{f}_\xi}{\exp\left(-\alpha C(X_t, U_t)\right)\u_h}}{\expect{\pib_{\thetab}, \hat{f}_\xi}{\exp\left(-\alpha C(X_t, U_t)\right)}} + 0\\
&= \frac{\expect{\pib_{\thetab}, \hat{f}_\xi}{\exp\left(-\alpha C(X_t, U_t)\right)\u_h}}{\expect{\pib_{\thetab}, \hat{f}_\xi}{\exp\left(-\alpha C(X_t, U_t)\right)}} \quad ,
\end{align}
recovering the MPPI update rule.

\textbf{Probability of Low Cost (PLC) \quad}\label{app:prob_low_cost} Similarly to \cite{dmd-mpc}, we can incorporate a threshold-utility to indicate preference for costs below a given threshold, using the indicator function: $\L(\tau) = \mathbbm{1}_{\cost \leq \cost_{t,\max}}(\cost(\xb_t, \ub_t))$. The likelihood then takes the form
\begin{align}
\expect{\pib_{\thetab}, \hat{f}_\xi}{\ \L(\tau)} &=  \expect{\pib_{\thetab}, \hat{f}_\xi}{\mathbbm{1}_{\cost \leq \cost_{t,\max}}(\cost(\xb_t, \ub_t))}.
\intertext{with the resulting gradient:}
\nabla_{\thetab} \log \expect{\pib_{\thetab}, \hat{f}_\xi}{\ \L(\tau)}  &= 
\frac{\expect{\pib_{\thetab}, \hat{f}_\xi}{\mathbbm{1}_{\cost \leq \cost_{t,\max}}(\cost(\xb_t, \ub_t))\nabla_\thetab\log  \pib_{\thetab}}}{\expect{\pib_{\thetab}, \hat{f}_\xi}{\mathbbm{1}_{\cost \leq \cost_{t,\max}}(\cost(\xb_t, \ub_t))}}
\end{align}
The threshold parameter $\cost_{t,\max}$ is set adaptively as the largest cost of the top member in the elite fraction of sampled trajectories. Using the same derivation for the case of Exponentiated Utility likelihood, the choice of a Gaussian policy with a threshold-utilty reduces to the update rule for the Cross Entropy Method~\cite{cem}:
\begin{align}
\theta'_h = \frac{\expect{\pib_{\thetab}, \hat{f}_\xi}{\mathbbm{1}_{\cost \leq \cost_{t,\max}}(\cost(\xb_t, \ub_t))\u_h}}{\expect{\pib_{\thetab}, \hat{f}_\xi}{\mathbbm{1}_{\cost \leq \cost_{t,\max}}(\cost(\xb_t, \ub_t))}} \quad .
\end{align}

\section{\textbf{Nonparametric Bayesian MPC}}\label{app:nonparam_mpc}

Inference can be performed directly over the posterior of control input sequences $\thetab \triangleq  \left( \u_t, \u_{t+1}, ..., \u_{t+H-1}\right)$. Assuming the likelihood gradient in the Bayesian MPC setting can be derived for differentiable dynamics:
\begin{align}
\nabla_{\thetab} \log \expect{\hat{f}_\xi}{\ \L(\xb_t)}  &= \frac{ \expect{\hat{f}_\xi}{\L(\xb_t)\, \nabla_\thetab \log p(\xb_t | \thetab; \xi, \x_t)}}{\expect{\hat{f}_\xi}{\L(\xb_t)}}
\end{align}

Similarly to the parametric formulation, the form of the cost-likelihood $\L$ will result in a particular update rule. The gradient can generally be evaluated by approximating the expectations with Monte Carlo sampling of state trajectories given the controlled stochastic dynamics., and evaluating the gradients on sampled trajectories. Under certain conditions, however, it may be evaluated in closed form. Such is the case for a Linear-Quadratic-Gaussian (LQG) system, for example~\cite{theodorou2012duality}.

\section{Motion Planning}
\label{app:motion_planning}
A special case can be considered for deterministic dynamics. We can define the stochastic dynamics using a dirac measure on the space of trajectories: $p(\xb_t | \thetab; \xi, \x_t) = \delta (X_t - \overline{X}_t)$, where $\overline{X}_t = F(\thetab, \xi, \x_t)$, and $F$ performs consecutive application of the deterministic dynamics $f$. Setting $\L(\xb_t, \thetab_t) = \exp(-\alpha C(\xb_t, \thetab_t))$, the gradient then reduces to :
\begin{align}
\nabla_{\thetab} \log \expect{\hat{f}_\xi}{\ \L(\xb_t, \thetab)}  &= \frac{\nabla_\thetab \int \exp\left(-\alpha C(X_t, \thetab)\right) \delta(\xb_t - \overline{X}_t) dX_t}{\int \exp\left(-\alpha C(X_t, \thetab)\right) \delta(\xb_t - \overline{X}_t)dX_t} \\
&= \frac{\nabla_\thetab \exp\left(-\alpha C(\overline{X}_t, \thetab)\right)}{\exp\left(-\alpha C(\overline{X}_t, \thetab)\right)} \\
&= \frac{-\alpha \exp\left(-\alpha C(\overline{X}_t, \thetab)\right) \nabla_\thetab C(\overline{X}_t, \thetab)}{\exp\left(-\alpha C(\overline{X}_t, \thetab)\right)} \\
&= -\alpha \nabla_\thetab C(\overline{X}_t, \thetab)
\end{align}

The gradient can then be evaluated in a straightforward manner via back-propagation on the cost function, through the dynamics. 

 \section{Complexity}
     The implementation of SV-MPC requires the computation of the kernel Gram-matrix $K(\theta^i, \theta^j) \ \forall\ i, j \in (1, ..., m)$ for the SVGD update. This requires an inner-product operation for all particle pairs, resulting in a computational complexity of $\mathcal{O}(m^2hd)$ (where $m$ : number of particles, $h$ : horizon, $d$ : control dimension). 
     However, by exploiting structured kernels for trajectories, the scaling with respect to the horizon can be removed by parallel computation of kernel factors. Evaluation of a kernel factor is then a constant-time operation ($\mathcal{O}(1)$), and the overall SV-MPC complexity reduces to $\mathcal{O}(m^2)$. In practice a relatively low order of particles is required: $m << 1\times10^3$. The core bottleneck is typically the generation of rollout trajectories from control samples during each iteration of MPC, which can be performed in parallel but is linear in time with respect to the horizon $h$. As with most MPC applications, the horizon length, number of samples, etc. can be varied to balance accuracy with runtime complexity, depending on the constraints of the system.

 \section{Experimental Details}
  
    The environment and controller parameters used in the experiments are summarized in tables \ref{table:env_param} and \ref{table:control_param}, respectively. Note for Table \ref{table:control_param}: the utility function parameter, such as $\alpha$ and \textit{elite fraction}, share the same values as those used by SV-MPC for either EU or PLC likelihood.
    
    \textbf{Planar navigation \quad}\label{app:exp_mpc_nav}
    The SV-MPC controller with exponentiated-utility (EU) is constructed with a set of 6 to 32 particles, where the gradient of each particle is estimated via Monte-Carlo sampling of control and state trajectories. At each round, the best-performing particle is chosen to generate the action using \eqref{eqn:sample_action}. 
    
    For state $\x_t = (x_t, \dot{x}_t)$ and control $\u_t = \Ddot{x}_t$, where $x_t$, $\dot{x}_t$, $\Ddot{x}_t$ are the 2D position, velocity and acceleration, respectively, and the $x_{goal}$ the target 2D goal position, we define the instantaneous and terminal costs (with respect to \eqref{eq:cost}):
    \begin{align}
        c(\x_t, \u_t) = 0.5(x_t-x_{goal})^\top(x_t-x_{goal}) + 0.25\dot{x}_t^\top\dot{x}_t + 0.2\Ddot{x}_t^\top\Ddot{x}_t\\
        c_{\mathrm{term}}(\x_t) = 1000(x_t-x_{goal})^\top(x_t-x_{goal}) + 0.1\dot{x}_t^\top\dot{x}_t
    \end{align}

    \textbf{Manipulation \quad}\label{app:exp_mpc_manip}  
    Leveraging parallel simulation provides the ability to efficiently compute highly-resolved geometric constraints between the robot and obstacles, eliminating the need for coarsely-defined heuristics for collision detection (such as signed-distance fields~\cite{mukadam2018continuous}). Although the dynamics are deterministic, the problem remains challenging since the posterior probability distribution implies a nonlinear mapping (via inverse kinematics) from the work-space to the sampling space.
    
    The cost-function consists of a cost on cartesian distance-to-goal from the end-effector, as well as a penalty on control.  For state $\x_t = (e_t, \dot{e}_t)$ and control $\u_t = \dot{q}_t$, where $e_t$, $\dot{e}_t$, are the cartesian end-effector positions and velocities, respectively, $\dot{q}$ the joint velocities, and the $e_{goal}$ the target 3D goal position, we define the instantaneous and terminal costs (with respect to \eqref{eq:cost}):
    \begin{align}
        c(\x_t, \u_t) = 1(e_t-e_{goal})^\top(e_t-e_{goal}) + 0.25\dot{e}_t^\top\dot{e}_t + 0.1\dot{q}_t^\top\dot{q}_t\\
        c_{\mathrm{term}}(\x_t) = 5000(e_t-e_{goal})^\top(e_t-e_{goal}) + 0.1\dot{e}_t^\top\dot{e}_t
       \end{align}    

    \textbf{Stochastic HalfCheetah \quad}\label{app:exp_halfcheetah}
    We modify the cost function to reward forward velocity only if the agent is forward-facing, as done in \cite{Okada2019VariationalIM}, along with a control penalty. Without this alteration, progress can be made fairly easily by applying torque commands in a single direction, and `cart-wheeling' the system. For instantaneous forward velocity $v_t$ and body angle $\beta$, 
    \begin{align}
        c(\x_t, \u_t) = 0.1\u_t^\top\u_t - \frac{v_t}{2}(1+\mathrm{sgn}(\cos \beta ))
    \end{align}       
    
     \textbf{Motion Planning \quad}\label{app:exp_plan}  The dynamics consist of a deterministic, velocity controlled single-integrator model on the 2D position: $\x_{t+1}=\x_t + \u_t\Delta t$, where $\x_t, \u_t \in \mathbb{R}^{d\times 1}$ ($d=2$). In the example, we use a constant-control (\textit{i.e.} zero-acceleration) prior on velocities: $\u_{t+1} = \u_t + \w_t,\  \w_t \sim \mathcal{N}(0, \Sigma)$. We can construct the prior over sequences by first considering the convolution of control inputs over the planning horizon $T$:
     \begin{align}
        \begin{bmatrix}
        \u_0 \\ \u_1 \\ \u_2 \\ \vdots \\ \u_{T-1}
        \end{bmatrix}
        = \begin{bmatrix} 
            I & 0 & 0 &\cdots & 0 \\
            I & I & 0 &\cdots & 0 \\
            I & I & I &\cdots & 0 \\
            \vdots & \vdots & \ddots & \vdots \\
            I & I & I &\cdots & I
        \end{bmatrix} \begin{bmatrix} \w_{-1} \\ \w_0 \\ \w_1 \\ \vdots \\ \w_{T-2} \end{bmatrix},
    \end{align}
    where we assume the initial control is drawn from the same zero-mean distribution: $\u_0 = \w_{-1}\sim \mathcal{N}(0, \Sigma)$. For the sequences $U\triangleq[\u_0, \u_1, ..., \u_{T-1}]^\top$ and $W\triangleq[\w_{-1},\w_0, \w_1, ..., \w_{T-2}]^\top$, we can write the above as:
    \begin{align}
        U = L W
    \end{align}
     with $L$ defined as the lower-triangular matrix. We can then define the prior over control sequences:
     \begin{align}
         U \sim p(U) = \mathcal{N}(\mathbf{0}, \mathbf{\Sigma})
     \end{align}
     with covariance $\mathbf{\Sigma}=LDL^T$, where $D=\mathrm{diag}(\Sigma, \Sigma, ..., \Sigma)\in \mathbb{R}^{Td \times Td}$. Note that the precision matrix $\mathbf{\Sigma}^{-1}$ is tri-diagonal, implying a graphical structure with Markovian dependencies. In the SV-MPC optimization, the prior can also be interpreted as a penalty which encourages smooth state trajectories (similarly to the GP-prior described in \cite{mukadam2018continuous}). In the motion planning problem, we use this multi-variate Gaussian as the prior over particles: $p(\thetab)=\mathcal{N}\left(0, \mathbf{\Sigma}\right)$. 
     
     Furthermore, the cost function defined for this problem includes a smooth obstacle cost-map. This is generated using a bi-modal mixture of Gaussians, with the probability of collision given by $p_{\mathrm{obs}}(x_t)$. The cost function is then:
    \begin{align}
        c(\x_t) &= 1\times10^5\, p_{\mathrm{obs}}(x_t)\\
        c_{\mathrm{term}}(\x_t) &= 1000(x_t-x_{goal})^\top(x_t-x_{goal})
    \end{align}

       \begin{table}[ht]
        \setlength\intextsep{0pt}
        \captionsetup{belowskip=0pt}
        \setlength{\belowcaptionskip}{-1pt}
        \centering
    	\caption{Environment Parameters}
    	\label{table:env_param}
    	\begin{tabular}{L{1.8cm}|*{6}{C{1.2cm}|}C{1.6cm}}
    	      Experiment & Episode length & Trials & state space $(\x_t \in)$ & control space $(\u_t \in)$ & 
    	      control limits & time-step ($\Delta t$)& dyn. noise $\sigma_{dyn}^2$\\ \hline \hline
    	    
        Planar Nav.     & 300 & 25 & $\mathbb{R}^4$ & $\mathbb{R}^2$ & $\left[-50, 50\right]$  & 0.015 & 0.1 \\ \hline
        Manipulation    & 150 & 24 & $\mathbb{R}^{14}$ & $\mathbb{R}^7$ & $\left[-1, 1\right]$ & 0.067 & ---   \\ \hline
        Stoch. HalfCheetah     & 300 & 16  & $\mathbb{R}^{18}$ & $\mathbb{R}^6$  &  $\left[-1, 1\right]$ & 0.05 & 0.25 \\ \hline
    	\end{tabular}
    \end{table}

    \begin{table}
        \setlength\intextsep{0pt}
        \captionsetup{belowskip=0pt}
        \setlength{\belowcaptionskip}{-1pt}
        \centering
    	\caption{Experiment Controller Settings.}
    	\label{table:control_param}
    	\begin{tabular}{p{1.5cm}|p{4.5cm}|*{2}{C{2.cm}|}C{2.cm}}
    	\multicolumn{2}{l|}{}                                      & Planar Nav. & Manipulation & Stoch. HalfCheetah  \\ \hline \hline
    	\multicolumn{2}{l|}{Planning Horizon ($H$)}                &  64        &      64      &   30     \\ 
    	\multicolumn{2}{l|}{Warm-start iterations}                 &  30        &      25       &  20   \\ 
    	\multicolumn{2}{l|}{Optimization iterations per timestep}  &  1         &      1       &    5   \\
    	\multicolumn{2}{l|}{Control variance  ($\sigma^2$)}        & 100        &     0.25      &  1   \\ \hline
    	\multirow{5}{*}{SV-MPC} & num. particles ($m$)             &  6, 12, 32 &      4       &   5  \\ 
    	                        & num. ctrl. samples per particle  &  8         &      32       &   32  \\ 
    	                        & step-size ($\epsilon$)           &  10        &      0.1        &   1  \\ 
    	                        & cost-likelihood ($\L(\cdot)$)      &  EU        &      EU       &  EU, PLC\\ \hline
    	\multirow{2}{*}{MPPI} & num. control samples               &  32        &       128     &  160  \\ 
                                & inverse-temperature ($\alpha$)   & $1\times10^{-3}$ &  0.1    &  1 \\ \hline
    	\multirow{2}{*}{CEM} & num. control samples                &  32        &       128      &  160  \\ 
                                & elite fraction                   &  0.1        &      0.1      &  0.1 \\ \hline
    	\end{tabular}
    \end{table}
        
 \section{Additional Results}
    \label{app:additional_results}
    \textbf{Planar Navigation \quad} Table~\ref{table:stein_vs_mpc}, we include quantitative summaries of performance across control types.  The performance of the SV-MPC controller improves with increasing particle number. 
    
    \begin{figure}[h]
    \setlength\intextsep{0pt}
    \captionsetup{belowskip=0pt}
    \setlength{\belowcaptionskip}{-1pt}
    \centering
	\caption{Statistics for planar navigation task over 25  trials (4x4 obstacle grid)}
	\label{table:stein_vs_mpc}
	\begin{tabular}{p{1.4cm}|p{1.1cm} V{3} p{1.4cm}|p{1.2cm}}
		Controller & Num. of particles & Avg. cost of success ($\times 10^3$) &  Success rate ($\%$) \\
		\specialrule{0.15em}{0em}{0em}

         & \centering 32 & \centering 20.7 & 96 \\\cline{2-4}
        SV-MPC & \centering 12 & \centering 21.8 & 96 \\\cline{2-4}
        & \centering 6 & \centering 25.6 &  84 \\ \cline{1-4}
        MPPI & \centering --- & \centering 26.5  &  64 \\ \cline{1-4}
        CEM & \centering --- & \centering 25.4  &  64\\
	\end{tabular}
    \end{figure}        
        
  \end{appendices}

\end{document}